\documentclass[10pt,journal,compsoc]{IEEEtran}
\usepackage{cite}
\usepackage{url}
\usepackage{ragged2e}
\usepackage{footnote}
\makesavenoteenv{tabular}
\makesavenoteenv{table}
\usepackage{epsfig}
\usepackage{graphicx}
\usepackage{amsmath,amssymb} 
\usepackage{algorithm}
\usepackage{algorithmicx}
\usepackage{algpseudocode}
\usepackage{graphics}
\usepackage{threeparttable}
\usepackage{color}
\usepackage[normalem]{ulem}
\usepackage{multirow}
\usepackage{float}
\usepackage{amsfonts}
\usepackage{bm}
\usepackage{array}
\usepackage{pifont}
\usepackage{diagbox}
\usepackage{rotating}
\usepackage{booktabs}
\usepackage{overpic}
\usepackage{textcomp}
\usepackage{contour}
\usepackage{makecell}

\usepackage{enumitem}
\usepackage{colortbl}
\usepackage[american]{babel}
\usepackage{microtype}
\usepackage{bbding}
\usepackage[breaklinks=true,colorlinks]{hyperref}

\newcommand{\figref}[1]{Fig.~\ref{#1}}
\newcommand{\tabref}[1]{Table~\ref{#1}}
\newcommand{\equref}[1]{Eq.~(\ref{#1})}
\newcommand{\secref}[1]{$\S$\ref{#1}}

\ifdefined \GramaCheck
  \newcommand{\CheckRmv}[1]{}
  \renewcommand{\eqref}[1]{Equation 1}
  \renewcommand{\equref}[1]{Equation 1}
  \renewcommand{\figref}[1]{Figure 1}
  \renewcommand{\tabref}[1]{Table 1}
\else
  \newcommand{\CheckRmv}[1]{#1}
  \renewcommand{\eqref}[1]{Equation~(\ref{#1})}
\fi

\usepackage{silence}
\graphicspath{{./Imgs/},{./Imgs/examples/},,{./Imgs/depth_examples/}{./Imgs/ErrorAnalysis/},{./Imgs/Visualization/},{./Imgs/authors/}}
\DeclareGraphicsExtensions{.pdf,.jpg,.png,.gif}

\def\methodname{{MobileSal}}
\def\sArt{{state-of-the-art~}}

\newcommand{\revise}[1]{{\textcolor[RGB]{0, 0, 0}{#1}}} 

\def\ie{\emph{i.e.}}

\def\etc{\emph{etc}}
\def\etal{{\em et al.~}}
\def\sArt{{state-of-the-art~}}

\begin{document}

\title{MobileSal: Extremely Efficient \\ RGB-D Salient Object Detection}

\markboth{IEEE TRANSACTIONS ON PATTERN ANALYSIS AND MACHINE INTELLIGENCE}%
{W\MakeLowercase{u} \MakeLowercase{\textit{et al.}} MobileSal: Extremely Efficient  RGB-D Salient Object Detection}

\author{
  Yu-Huan Wu, Yun Liu, Jun Xu, Jia-Wang Bian, Yu-Chao Gu, and Ming-Ming Cheng
  \IEEEcompsocitemizethanks{
      \IEEEcompsocthanksitem Corresponding author: M.-M. Cheng. (E-mail: cmm@nankai.edu.cn)
    \IEEEcompsocthanksitem Y.-H.~Wu,  Y. Liu, J.~Xu, Y.-C. Gu, and M.-M.~Cheng 
      are with the TKLNDST, College of Computer Science, 
      Nankai University, Tianjin, China, 300350. 
    \IEEEcompsocthanksitem J.-W.~Bian is with the University of Adelaide.
    }
}

\IEEEtitleabstractindextext{%
\begin{abstract}
  \justifying
The high computational cost of neural networks has prevented 
recent successes in RGB-D salient object detection (SOD) 
from benefiting real-world applications.
Hence, this paper introduces a novel network, \methodname, 
which focuses on efficient RGB-D SOD using 
mobile networks for deep feature extraction.
However, mobile networks are less powerful in feature representation
than cumbersome networks.
To this end, we observe that the depth information of color images can strengthen 
the feature representation related to SOD if leveraged properly.
Therefore, we propose an implicit depth restoration (IDR) technique to 
strengthen the mobile networks' feature representation capability for RGB-D SOD.
IDR is only adopted in the training phase and is omitted
during testing, so it is computationally free.
Besides, we propose compact pyramid refinement (CPR) for efficient multi-level
feature aggregation to derive salient objects with clear boundaries.
With IDR and CPR incorporated, 
\methodname~performs favorably against \sArt  methods
on six challenging RGB-D SOD datasets with much faster speed 
(450fps for the input size of $320\times 320$)
and fewer parameters (6.5M).
The code is released at \url{https://mmcheng.net/mobilesal}.
\end{abstract}

}

\maketitle

\IEEEdisplaynontitleabstractindextext

\IEEEpeerreviewmaketitle

\section{Introduction} \label{sec:intro}
Salient object detection (SOD) aims to locate and segment 
the most eye-catching object(s) in natural images. 
It is a fundamental problem in image understanding and serves as a preliminary step 
for many computer vision tasks such as 
visual tracking~\cite{hong2015online},
content-aware image editing~\cite{wang2018deep},
and weakly supervised learning~\cite{liu2020leveraging}.
Current SOD methods are mainly developed for RGB images \cite{zhang2017amulet,pang2020multi,zhao2020suppress},
which are usually hindered by indistinguishable foreground and background textures. 
To this end, researchers resort to the easily accessible depth information as 
an important complement to the RGB counterpart, with promising progress in RGB-D SOD
\cite{chen2019three,zhao2019constrast,piao2019depth,fan2020rethinking,zhang2020uc,zhao2020single,li2020rgb,fu2020jl}.

\newcommand{\AddImg}[1]{\includegraphics[width=.47\textwidth]{#1}}

\CheckRmv{
\begin{figure}[!t]
\centering
\renewcommand{\arraystretch}{0.5}
  \setlength{\tabcolsep}{0.3mm}
\begin{tabular}{c}
\AddImg{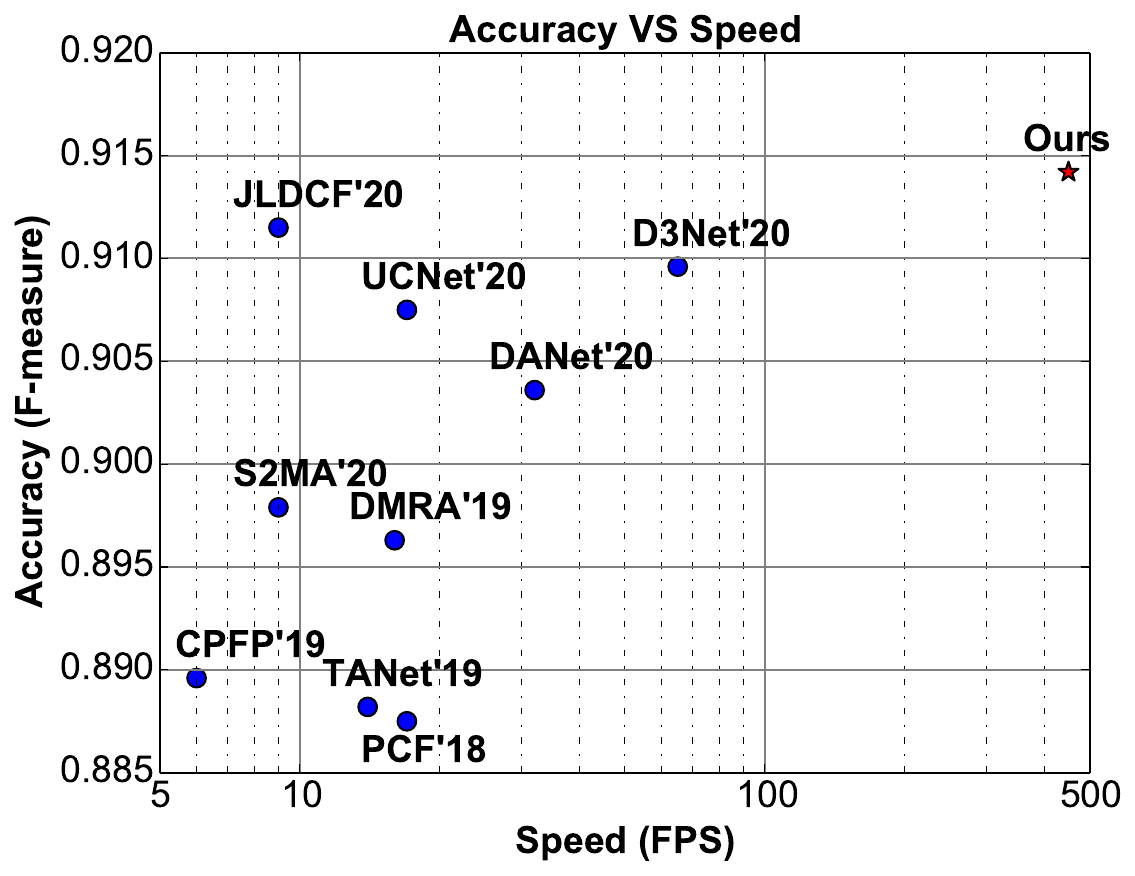}
\\
\end{tabular}
\caption{
Comparison with \sArt methods (see references in \tabref{tab:benchmark}) on the challenging NJU2K~\cite{ju2014depth} dataset. Our method (MobileSal shows very competitive accuracy and much faster speed.}\label{fig:speed_acc}
\end{figure}
}

While convolutional neural networks (CNNs) have made brilliant achievements on RGB-D SOD
\cite{zhao2019constrast,piao2019depth,fan2020rethinking,zhang2020uc,zhao2020single,fu2020jl},
their high accuracy often comes at the expense of high computational costs and large model size.
This situation has prevented recent \sArt methods~\cite{zhao2019constrast, fu2020jl} from being applied to real-world applications,
especially for those on mobile devices, 
which are depth accessible,
with very limited energy overhead and computational capability.
Hence, it is essential to design efficient networks for accurate RGB-D SOD.
A na\"ive solution towards this goal is to adopt lightweight backbones such as 
MobileNets \cite{howard2017mobilenets,sandler2018mobilenetv2} 
and ShuffleNets \cite{zhang2018shufflenet,ma2018shufflenet} for deep feature extraction, 
instead of commonly-used cumbersome backbones like VGG~\cite{simonyan2014very} and ResNets~\cite{he2016deep}.
The problem is that lightweight networks are usually less powerful than cumbersome networks 
on feature representation learning, as widely acknowledged by the research community 
\cite{howard2017mobilenets,sandler2018mobilenetv2,zhang2018shufflenet,ma2018shufflenet}.
This problem would hinder lightweight networks from accurate RGB-D SOD performance.

To overcome this challenge, we note that the depth information of color images, if leveraged properly, can strengthen the feature representation for RGB-D SOD~\cite{zhao2019constrast,piao2019depth}.
Unlike some existing studies~\cite{zhao2020single, zhang2020uc} that leverage the depth information explicitly, 
in this paper, we propose an implicit depth restoration (IDR) technique 
to strengthen the feature 
representation learning of the lightweight backbone network so as to ensure the accuracy
of RGB-D SOD in an efficient setting.
More importantly, IDR is only adopted in the training phase and is omitted during testing,
so it is \textit{computationally free} during the inference stage.
Specifically, we enforce our model to restore the depth map from high-level backbone features,
through which the representation learning of the lightweight backbone becomes more powerful
with important supervision on the depth stream.
Besides the IDR module, we propose two more components to ensure
the high efficiency: i) we conduct RGB and depth information
fusion only at the coarsest level, because such a small
feature resolution (\ie, 1/32 scale) is essential for reducing
computational cost; ii) we propose a compact pyramid
refinement (CPR) module to efficiently aggregate multi-scale
deep features, for accurate SOD with clear boundaries.

With MobileNetV2 \cite{sandler2018mobilenetv2} as the backbone network,
\methodname~achieves 450fps on a single NVIDIA RTX 2080Ti GPU with the input size of $320\times 320$,
tens of times faster than existing RGB-D SOD methods~\cite{zhao2019constrast, piao2019depth, liu2020learning, fu2020jl}.
Extensive experiments on six challenging datasets demonstrate that 
\methodname~also achieves competitive performance compared with \sArt methods
(max F-measure of 91.4\% and 91.2\% on NJU2K~\cite{ju2014depth} and 
DUTLF~\cite{piao2019depth} datasets, respectively)
with fewer parameters (6.5M).
Such high efficiency, good accuracy, and small model size would benefit
many real-world applications.

In summary, our main contributions include:
\begin{itemize}
\item To the best of our knowledge, \methodname~is the first to shed light upon
efficient RGB-D SOD by proposing an extremely efficient network with a speed of 450fps.
\item To ensure the efficiency of \methodname~on cross-modal fusion, 
\methodname~fuses RGB 
and depth information only at the coarsest level and then efficiently aggregates 
multi-level deep features using a compact pyramid refinement (CPR) module.
\item To ensure the accuracy of \methodname, we propose an implicit depth 
restoration (IDR) technique to strengthen the less powerful features learned 
by mobile backbone networks. 
This technique is also applicable to other segmentation tasks such as RGB-D semantic segmentation.
\end{itemize}

\CheckRmv{
\begin{figure*}[!t]
    \centering
    \includegraphics[width=\textwidth]{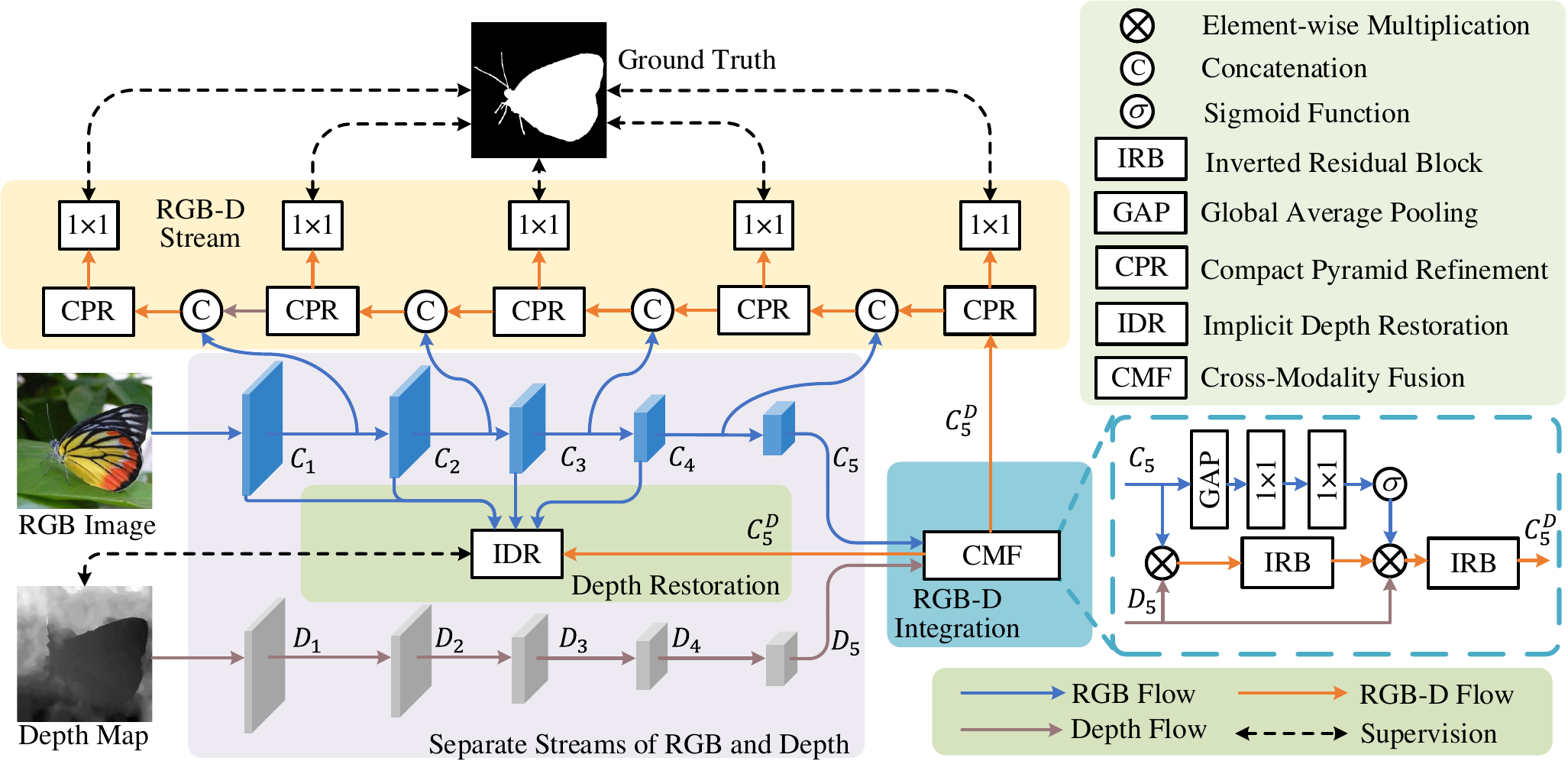}
    \caption{
    \textbf{The pipeline of MobileSal}.
    We fuse RGB and depth information only at the coarsest level
    and then efficiently do the multi-scale aggregation with CPRs.
    The IDR branch strengthens the less powerful features learned by the mobile networks in a computationally free manner.
}\label{fig:pipeline}
\end{figure*}
}

\section{Related Work} 
\subsection{Salient Object Detection}
Benefited from the rapid development of deep CNNs 
\cite{simonyan2014very,he2016deep} in recent years,
CNNs-based SOD methods for RGB images
\cite{hou2019deeply,qin2019basnet,liu2019pool,pang2020multi,zhang2019salient,zhao2020suppress,wu2021regularized,wu2021edn} 
have achieved substantial progress compared with conventional methods
\cite{yang2013saliency,cheng2013efficient,cheng2014global,wang2017salient,li2013saliency,zhang2017saliency}.
Along this direction, much attention is paid to design various effective strategies to fuse multi-scale features
generated by multi-level CNN layers
\cite{hou2019deeply,qin2019basnet,zhang2019salient,zhao2020suppress}.
Some efforts are also spent on exploring the effectiveness of extra boundary information
\cite{liu2019pool,qin2019basnet,wu2019mutual} or part-object relationship \cite{liu2021part}.
Detailed introductions of SOD works can refer to recent popular surveys \cite{han2018advanced, borji2019salient, wang2021salient}.
Despite many success stories, RGB SOD is hindered by indistinguishable foreground and background textures,
which can be largely alleviated by incorporating the depth information, \ie, RGB-D SOD.

\subsection{RGB-D Salient Object Detection} \label{sec:rgb-d}
Like early SOD methods, conventional RGB-D SOD works extract hand-crafted features from RGB and depth maps and fuse them together
\cite{lang2012depth, ciptadi2013depth, desingh2013depth, fan2014salient, peng2014rgbd, cheng2014depth}.
Recently, RGB-D SOD has gained more attention and 
deep-learning-based RGB-D SOD has been developed rapidly
\cite{zhang2020uc,li2020rgb,zhao2020single,zhao2019constrast,piao2019depth, fan2020rethinking, pang2020hierarchical, zhao2021self, chen2020progressively, huang2020joint, qu2017rgbd}.
Typically, Zhao \etal \cite{zhao2019constrast} proposed a contrast-prior-based network, providing strong depth enhancement.
Piao \etal \cite{piao2019depth} proposed to refine depth via depth-induced 
multi-scale recurrent attention (DMRA).
Huang \etal \cite{huang2020joint} proposed RGB-D fusion via joint cross-modal and unimodal features, providing more comprehensive RGB-D analysis.
Zhang \etal \cite{zhang2020uc} provided another perspective of uncertainty RGB-D saliency via conditional variational autoencoders.
Chen \etal \cite{chen2021rgb} first introduced 3D CNNs to RGB-D SOD, providing more abundant spatial semantics.
Ji \etal \cite{ji2021calibrated} introduced a flexible depth calibration module, providing reliable complementary information for saliency models.
Zhao \etal \cite{zhao2021self} proposed a self-supervised learning framework, which only leverages image-level annotations saving large costs from large-scale data annotations.
More inspiring related works can refer to the recent survey \cite{zhou2021rgb}.

In terms of the fusion strategy of RGB and depth information,
the diagrams of RGB-D SOD architectures can be broadly divided into 
late fusion~\cite{desingh2013depth, han2017cnns, wang2019adaptive},
early fusion \cite{zhang2020uc,li2020rgb,zhao2020single, chen2021rgb}, 
and multi-scale fusion
\cite{zhao2019constrast,piao2019depth,fu2020jl,liu2020learning,pang2020hierarchical, zhao2021self, chen2020progressively, huang2020joint}.
Late fusion appears at the end of feature extraction
and only predicts the result from the fused features \cite{desingh2013depth,han2017cnns, wang2019adaptive}.
Early fusion directly concatenates the input RGB image and depth map and then derives the saliency map
from such RGB-D input using the encoder-decoder network \cite{zhao2020single} or hypercolumn network \cite{zhang2020uc}.
Multi-scale fusion first extracts RGB and depth features 
separately and then aggregates RGB-D features at all levels \cite{piao2019depth, huang2020joint},
at middle and high levels \cite{huang2021employing},
or at middle levels \cite{pang2020hierarchical}.
Although the early fusion strategy is more efficient, multi-scale fusion is more accurate.
To ensure high efficiency, our method only fuses RGB and depth features at the coarsest level
in a small resolution.
IDR is then applied to strengthen the feature representation learning of 
mobile networks in a computationally free manner.

\subsection{Efficient Backbone Networks}
Recent growing interests in mobile vision applications have generated a high demand for 
efficient CNNs.
Mobile devices such as autonomous driving vehicles, robots, and smartphones only have limited 
computational resources, so traditional cumbersome networks, like VGG~\cite{simonyan2014very} and
ResNets~\cite{he2016deep}, 
are unsuitable for these platforms.
To this end, some efficient networks are proposed for image classification,
such as MobileNets \cite{sandler2018mobilenetv2}, 
ShuffleNets \cite{ma2018shufflenet}, 
MnasNet~\cite{tan2019mnasnet},~\etc.
There also emerge some efficient networks for 
semantic segmentation \cite{mehta2018espnetv2},
object detection \cite{tan2020efficientdet},
and ordinary RGB SOD \cite{liu2020lightweight, liu2021samnet}.
These efficient networks are with low computational costs and thus flexible for mobile platforms.
In this paper, we are the first to shed light upon efficient RGB-D SOD by adopting 
MobileNetV2~\cite{sandler2018mobilenetv2} as the backbone for deep feature extraction.
Our proposed techniques aim to ensure the SOD accuracy and high efficiency simultaneously
in such a lightweight setting.

\section{Methodology} \label{sec:method}
In this section, we first provide an overview of our method in \S\ref{sec:pipeline}. 
Then, we introduce the proposed cross-modal feature fusion scheme in \S\ref{sec:rgbd-fusion},
implicit depth restoration in \S\ref{sec:IDR},
compact pyramid refinement in \S\ref{sec:cpr}.
Finally, we present
the hybrid loss function in \S\ref{sec:loss}.

\subsection{Overview} \label{sec:pipeline}
\figref{fig:pipeline} depicts the overall architecture of our method.
We use RGB and depth streams for separate feature extraction.

\textbf{RGB Stream}.
We employ MobileNetV2~\cite{sandler2018mobilenetv2} as the backbone of our method.
To adapt it to the SOD task, we remove the global average pooling layer and 
the last fully-connected layer from the backbone.
For the RGB stream, each stage is followed by a convolutional layer with a stride of 2, 
and thus feature maps are downsampled into half resolution after each stage.
For convenience, we denote the output feature maps for five stages as
$\mathcal{C}_1,\mathcal{C}_2,\mathcal{C}_3,\mathcal{C}_4,\mathcal{C}_5$,
with strides of $2,2^2,2^3,2^4,2^5$, respectively.

\textbf{Depth Stream}.
Similar to the RGB stream, the depth stream also has five stages
with the same strides.
Since depth maps contain less semantic information than the corresponding RGB images,
we build a lightweight depth network with fewer convolutional blocks than the RGB stream.
Each stage only has two \textbf{Inverted Residual Blocks (IRB)} \cite{sandler2018mobilenetv2}.
Such a design reduces computational complexity that accords with the goal of efficient RGB-D SOD.
In each IRB, we first expand the feature map along the channel dimension 
by $M$ times via a $1\times 1$ convolution, followed by a depthwise separable $3\times 3$ convolution
\cite{howard2017mobilenets} with the same number of input and output channels.
Then, the feature channels are squeezed to $1/M$ via another $1\times 1$ convolution.
Here, each convolution is followed by Batch Normalization (BN)~\cite{ioffe2015batch} and 
ReLU~\cite{nair2010rectified} layers, except for the last $1\times 1$ convolution that only has a BN layer.
The final output of the inverted residual block is the element-wise sum of the initial input 
and the output generated by the above three sequential convolutions.
For the first layer in each stage, the stride of the depthwise separable convolution is set as 2,
and the number of hidden feature channels is increased if needed.
The output feature maps of five stages of the depth stream are denoted as 
$\mathcal{D}_1,\mathcal{D}_2,\mathcal{D}_3,\mathcal{D}_4,\mathcal{D}_5$, 
the first four of which have $16, 32, 64, 96$ channels, respectively.
$\mathcal{D}_5$ and $\mathcal{C}_5$ have the same number of channels and the same stride.

As shown in \figref{fig:pipeline}, with the outputs of the RGB and depth stream, 
we first fuse the extracted RGB feature $\mathcal{C}_5$ and depth feature $\mathcal{D}_5$ 
to generate the RGB-D feature $\mathcal{C}_5^D$.
The proposed IDR technique restores the depth map from 
$\mathcal{C}_1,\mathcal{C}_2,\mathcal{C}_3,\mathcal{C}_4,\mathcal{C}_5^D$,
which is supervised by the input depth map to strengthen the feature representation learning.
For saliency prediction, we design a lightweight decoder using the CPR module as the basic unit.
The output of the decoder at the bottom stage is the final predicted saliency map.
More details can be seen in the following sections.

\CheckRmv{
\begin{figure*}[!t]
    \centering
    \includegraphics[width=\textwidth]{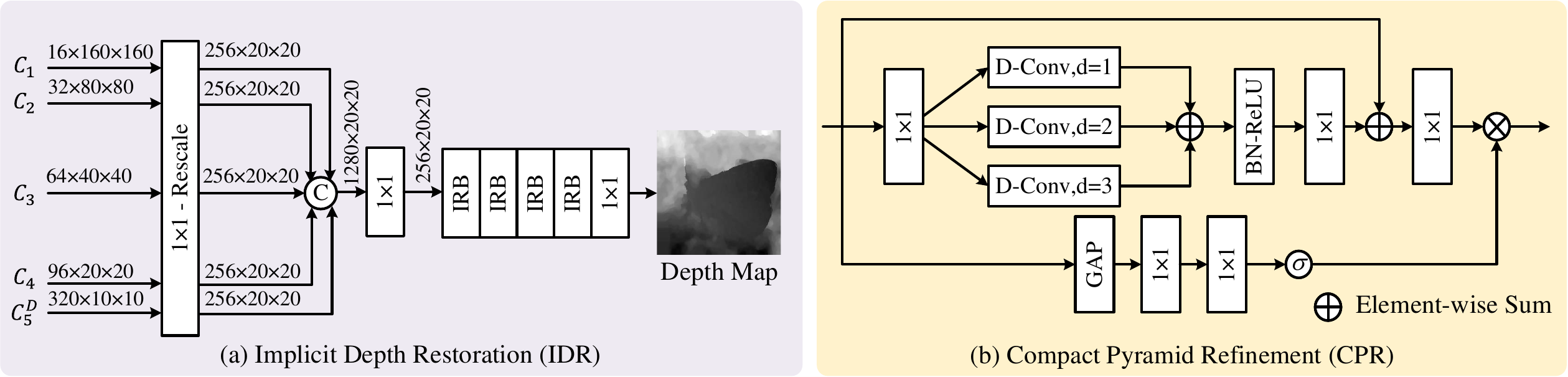}
    \vspace{-22pt}
    \caption{\textbf{Illustration of the proposed IDR and CPR.} 
    (a) The IDR branch strengthens the less powerful features of the mobile backbone network.
    (b) Multi-level deep features are efficiently aggregated by the CPR module.
    ``D-Conv'' indicates depthwise separable convolution.
} \label{fig:IDR_cpr}
\end{figure*}
}

\subsection{Cross-Modal Fusion of RGB and Depth Features}\label{sec:rgbd-fusion}
The depth map reveals spatial cues of color images, which helps distinguish
the foreground objects from the background, especially for scenarios with complicated textures.
As demonstrated by previous studies
\cite{zhao2019constrast,piao2019depth,fu2020jl,liu2020learning,pang2020hierarchical,chen2020progressively},
proper RGB and depth feature fusion is essential for accurate RGB-D SOD.
Our main consideration here is to ensure the high efficiency of our method.
Instead of conducting fusion at multiple levels
\cite{zhao2019constrast,piao2019depth,fu2020jl,liu2020learning,pang2020hierarchical,chen2020progressively, wu2021jcs},
we only fuse RGB and depth features at the coarsest level because the small feature resolution
leads to low computational cost.

According to the above analyses, we only fuse the RGB feature map $\mathcal{C}_5$ and 
the depth feature map $\mathcal{D}_5$.
We design a lightweight \textbf{Cross-Modal Fusion (CMF)} module for this purpose,
as shown in \figref{fig:pipeline}.
Intuitively, semantic information mainly exists in the RGB image.
The depth map conveys the prior of depth-smooth regions that 
approximately represent the shapes and structures of complete 
objects or stuff.
Hence, we adopt depth features like a gate to enhance 
RGB semantic features through multiplication,
which can be viewed as a strong regularization.
Note that element-wise addition or concatenation can only
aggregate two feature maps by treating features equally,
which is orthogonal to our goal.
Experiments in \S\ref{sec:ablation} also demonstrate our hypothesis.

Specifically, we first combine the RGB and depth features with an above-mentioned IRB to derive 
the transited RGB-D feature maps $\mathcal{T}$, which can be formulated as 
\begin{equation}\label{equ:initial_rgbd_fuse}
  \mathcal{T} ={\rm IRB}(\mathcal{C}_5 \otimes \mathcal{D}_5),
\end{equation}
where and $\otimes$ is the element-wise multiplication operator.
Meanwhile, we apply a global average pooling (GAP) layer to $\mathcal{C}_5$ to get 
a feature vector, followed by two fully-connected layers to 
compute the RGB attention vector $\mathbf{v}$, like
\begin{equation}\label{equ:catt}
  \mathbf{v} = \sigma({\rm FC}_2({\rm ReLU}({\rm FC}_1({\rm GAP}(\mathcal{C}_5))))),
\end{equation}
in which ${\rm FC}$ and ${\rm ReLU}$ denote fully-connected and ReLU layers, respectively.
The number of output channels of ${\rm FC}_1$ and ${\rm FC}_2$ is the same as the input.
$\sigma$ indicates the standard sigmoid function.
With $\mathcal{T}$ and $\mathbf{v}$ computed, the multiplication of $\mathbf{v}$, $\mathcal{T}$, 
and $\mathcal{D}_5$ are fed into an IRB, like
\begin{equation}\label{equ:CMF}
  \mathcal{C}_5^D ={\rm IRB}(\mathbf{v}\otimes \mathcal{T} \otimes \mathcal{D}_5),
\end{equation}
where $\mathcal{C}_5^D$ indicates the output feature map of the CMF module.
Note that $\mathbf{v}$ is replicated to the same shape as $\mathcal{T}$ before multiplication.
\equref{equ:CMF} filters RGB semantic features again by multiplying $\mathcal{D}_5$,
and the channel attention $\mathbf{v}$ is used to recalibrate the fused features.
After the fusion of RGB and depth features, 
we can derive the backbone features, 
including the RGB features $\mathcal{C}_1,\mathcal{C}_2,\mathcal{C}_3,\mathcal{C}_4$,
and the fused RGB-D feature $\mathcal{C}_5^D$.

\subsection{Implicit Depth Restoration} \label{sec:IDR}
As widely acknowledged
\cite{howard2017mobilenets,sandler2018mobilenetv2,zhang2018shufflenet,ma2018shufflenet},
lightweight backbone networks are less powerful in feature representation learning
than cumbersome networks.
To ensure the accuracy of RGB-D SOD, we consider strengthening the representation learning
of mobile networks.
We observe that the depth map conveys depth-smooth regions that usually represent 
objects, object parts, or smooth background, because intuitively, 
an object or a connected stuff region usually has similar depth.
This observation motivates us to use the depth map as an extra supervision source to
guide the representation learning, which would help mobile networks restrain 
the texture changes within objects or connected stuff regions and highlight 
the difference among them.
In this way, the contrast between salient objects and the background will be strengthened too.
With this idea, we design an \textbf{Implicit Depth Restoration (IDR)} technique.
Here, we use the word ``implicit'' because IDR is only adopted in the training phase
and is omitted during testing, making it computationally free for practical deployment.

We continue by introducing how to use $\mathcal{C}_1,\mathcal{C}_2,\mathcal{C}_3,\mathcal{C}_4,\mathcal{C}_5^D$
for the above auxiliary supervision.
As shown in \figref{fig:IDR_cpr} (a), the pipeline of IDR is simple, \ie, 
just concatenating multi-level feature maps and then fusing them.
Specifically, we first apply a $1\times 1$ convolution to squeeze 
$\mathcal{C}_1,\mathcal{C}_2,\mathcal{C}_3,\mathcal{C}_4,\mathcal{C}_5^D$
to the same number of channels, \ie, 256.
Then, the resulting feature maps are resized to the same size as $\mathcal{C}_4$,
followed by the concatenation of them.
A $1\times 1$ convolution changes the concatenated feature map from 1280 channels
to 256 channels for saving computational cost.
Next, four sequential IRBs are followed to fuse multi-level features so that
we can obtain powerful multi-scale features.
At last, a simple $1\times 1$ convolution converts the fused feature map to a single channel.
With a standard sigmoid function and bilinear upsampling, we can obtain the restored 
depth map with the same size as the input.
The training loss of IDR adopts the well-known SSIM metric~\cite{wang2004image} to 
measure the structural similarity between the restored depth map $\mathcal{D}_r$ 
and input one $\mathcal{D}_g$, which can be written as 
\begin{equation}
    \mathcal{L}_{\rm IDR} = 1 -{\rm SSIM}(\mathcal{D}_r, \mathcal{D}_g),
\end{equation}
where SSIM uses the default setting.
Note that the above operations are omitted during testing to make IDR free.

\subsection{Compact Pyramid Refinement}\label{sec:cpr}
It is widely accepted that high-level features in the backbone network contain 
semantic abstract features, while low-level features convey fine-grained details.
For accurate SOD, it is essential to fully utilize both high-level and low-level features.
There exists a lot of literature on this topic
\cite{zhao2019constrast,piao2019depth,fu2020jl,liu2020learning,li2020rgb},
but existing methods usually design cumbersome decoders without consideration of efficiency.
Here, our decoder should not only fuse multi-level features effectively 
but also be efficient as much as possible.

The proposed decoder uses the \textbf{Compact Pyramid Refinement (CPR)} module as the basic unit.
For efficiency, CPR uses $1\times 1$ and depthwise separable convolutions \cite{howard2017mobilenets}
instead of vanilla convolutions in previous methods 
\cite{fu2020jl,liu2020learning,li2020rgb,zhao2020single}.
Since multi-level features exhibit multi-scale representations with the high level corresponding to 
the coarse scale and vice versa, multi-scale learning would be necessary for multi-level feature fusion.
Hence, CPR adopts a lightweight multi-scale learning strategy to enhance such fusion.
Suppose that the input of a CPR module is $\mathcal{X}$.
As shown in \figref{fig:IDR_cpr} (b), CPR first applies a $1\times 1$ convolution
to expand the number of channels by $M$ times.
Then, three $3\times 3$ depthwise separable convolutions with dilation rates of $1,2,3$ 
are connected parallel for multi-scale fusion.
This can be formulated as
\begin{equation}
\begin{aligned}
    \mathcal{X}_1\ &={\rm Conv}_{1\times 1}(\mathcal{X}),\\
    \mathcal{X}_2^{d_1} &={\rm Conv}_{3\times 3}^{d_1}(\mathcal{X}_1),\\
    \mathcal{X}_2^{d_2} &={\rm Conv}_{3\times 3}^{d_2}(\mathcal{X}_1),\\
    \mathcal{X}_2^{d_3} &={\rm Conv}_{3\times 3}^{d_3}(\mathcal{X}_1),\\
    \mathcal{X}_2\ &={\rm ReLU}({\rm BN}(\mathcal{X}_2^{d_1} + \mathcal{X}_2^{d_2} + \mathcal{X}_2^{d_3})),
\end{aligned}
\end{equation}
where $d_1$, $d_2$, and $d_3$ are dilation rates, \ie, $1,2,3$ here, respectively.
${\rm BN}$ is the abbreviation of batch normalization \cite{ioffe2015batch}.
A $1\times 1$ convolution is used to squeeze channels to the same number as the input, \ie,
\begin{equation}
    \mathcal{X}_3 ={\rm Conv}_{1\times 1}(\mathcal{X}_2) + \mathcal{X},
\end{equation}
which uses a residual connection for better optimization.
The attention mechanism in \equref{equ:catt} is applied to $\mathcal{X}$ 
to calculate an attention vector $\mathbf{v}'$, so that we have
\begin{equation}\label{equ:cpr}
    \mathcal{Y} = \mathbf{v}' \otimes{\rm Conv}_{1\times 1}(\mathcal{X}_3).
\end{equation}
\equref{equ:cpr} uses global contextual information to recalibrate the fused features.

As shown in \figref{fig:pipeline}, at each decoder stage, two feature maps from 
the top decoder and the corresponding encoder stage first reduce their numbers 
of channels to half using a $1\times 1$ convolution separately.
The results are then concatenated, followed by a CPR module for feature fusion.
In this way, our lightweight decoder aggregates multi-level features from top to bottom.

\subsection{Hybrid Loss Function}\label{sec:loss}
At each decoder stage, we predict the saliency map by sequentially adding
a $1\times 1$ convolution with a single channel, a sigmoid function, and bilinear upsampling 
to the output of the CPR module, as shown in \figref{fig:pipeline}.
Hence, we can derive predicted saliency maps $\mathcal{P}_i (i=1,2,\cdots,5)$
for five stages, respectively.
Let the ground-truth saliency map be $\mathcal{G}$. 
The loss of each side-output can be computed as 
\begin{equation}
  \mathcal{L}_{sal}^i ={\rm BCE}(\mathcal{P}_i, \mathcal{G}) +{\rm Dice}(\mathcal{P}_i, \mathcal{G}),
\end{equation}
${\rm BCE}$ denotes binary cross-entropy loss function:
\begin{equation}
  {\rm BCE}(\mathcal{P}_i, \mathcal{G})\ = \mathcal{G}\cdot\log \mathcal{P}_i + (1 - \mathcal{G})\cdot\log (1-\mathcal{P}_i),
\end{equation}
where ``$\cdot$'' indicates the dot-product operation.
${\rm Dice}$ represents the Dice loss \cite{milletari2016v}:
\begin{equation}
  {\rm Dice}(\mathcal{P}_i, \mathcal{G}) = 1 - \frac{2 \cdot \mathcal{G}\cdot \mathcal{P}_i}{||\mathcal{G}|| + ||\mathcal{P}_i||},
\end{equation}
where $||\cdot||$ denotes the $\ell_{1}$ norm.
With deep supervision and IDR, the training loss can be formulated as 
\begin{equation}\label{equ:loss_combine}
  \mathcal{L} = \sum_{i=1}^5 \mathcal{L}_{sal}^i + \lambda\cdot \mathcal{L}_{\rm IDR},
\end{equation}
where $\lambda$ is a balance weight.
In the testing phase, $\mathcal{P}_1$ is the final predicted saliency map.

\newcommand{\tabincell}[2]{\begin{tabular}{@{}#1@{}}#2\end{tabular}}

\begin{table*}[!t]
  \centering
  \small
  \caption{\textbf{Quantitative results on six challenging datasets}.
    The best, second best, and third best results are highlighted in
    \textcolor[rgb]{1,  0,  0}{\textbf{red}}, \textcolor[rgb]{0,  .69,  .941}{\textbf{blue}} 
    and \textbf{bold}, respectively.
    Our method achieves the best speed-accuracy trade-off.}
    \label{tab:benchmark}
  \renewcommand{\arraystretch}{1.04}
  \setlength\tabcolsep{1pt}
   \resizebox{\textwidth}{!}{
    \begin{tabular}{c|r|cccccccccccccccc}
      \Xhline{1pt}
      \multicolumn{2}{c|}{Method} & DESM  & LHM   & ACSD  & DCMC  & CTMF  & PCF   & TANet & CPFP  & DMRA  & D3Net & JLDCF & S2MA  & UCNet & DANet & BiANet & MobileSal \\
      \multicolumn{2}{c|}{\#PubYear [Ref]} &  2014 \cite{cheng2014depth}     &   2014 \cite{peng2014rgbd}   &    2014 \cite{ju2014depth}  &  2016 \cite{cong2016saliency}    &   2017 \cite{han2017cnns}     &   2018 \cite{chen2018progressively}   &   2019 \cite{chen2019three}    &   2019 \cite{zhao2019constrast}    &  2019 \cite{piao2019depth}    &   2020 \cite{fan2020rethinking}   &  2020 \cite{fu2020jl}    &  2020 \cite{liu2020learning}   &   2020 \cite{zhang2020uc}    &  2020  \cite{zhao2020single} & 2021 \cite{zhang2021bilateral}   & (Ours) \\
      \Xhline{1pt}
    \rowcolor[rgb]{0.9,0.9,0.9}
    \multicolumn{2}{c|}{Params (M)} & -      & -     & -     & -     & -     & 133.4  & 232.4  & 69.5  & 59.7  & 43.2  & 137.0  & 86.7  & \textbf{33.3}  & \textcolor[rgb]{0,  .69,  .941}{\textbf{26.7}} & 49.6  & \textcolor[rgb]{1,  0,  0}{\textbf{6.5}} \\
    \rowcolor[rgb]{0.9,0.9,0.9}
    \multicolumn{2}{c|}{Speed (FPS)} & -     & -     & 1     & -     & 8    & 17    & 14    & 6     & 16    & \textcolor[rgb]{0,  .69,  .941}{\textbf{65}} & 9     & 9     & 17    & 32 & \textbf{50} &  \textcolor[rgb]{1,  0,  0}{\textbf{450}} \\
    \cline{1-18}
    \multirow{5}[0]{*}{NJU2K} & $F_\beta^{\text{max}}$ $\uparrow$ &  0.767  & 0.703  & 0.749  & 0.759  & 0.857  & 0.887  & 0.888  & 0.890  & 0.896  & \textbf{0.910} & \textcolor[rgb]{ 0,  .69,  .941}{\textbf{0.912}} & 0.898  & 0.908  & 0.904  & 0.908  & \textcolor[rgb]{ 1,  0,  0}{\textbf{0.914}} \\
          & MAE $\downarrow$   & 0.286  & 0.204  & 0.200  & 0.171  & 0.085  & 0.059  & 0.060  & 0.053  & 0.051  & 0.047  & \textcolor[rgb]{ 1,  0,  0}{\textbf{0.041}} & 0.054  & \textbf{0.043} & 0.047  & 0.044  & \textcolor[rgb]{ 1,  0,  0}{\textbf{0.041}} \\
          & $S_\alpha$ $\uparrow$ & 0.671  & 0.515  & 0.708  & 0.686  & 0.849  & 0.877  & 0.878  & 0.878  & 0.886  & 0.900  & \textbf{0.902} & 0.894  & 0.897  & 0.897  & \textcolor[rgb]{ 0,  .69,  .941}{\textbf{0.904}} & \textcolor[rgb]{ 1,  0,  0}{\textbf{0.905}} \\
          & $E_\xi^{\text{max}}$ $\uparrow$ & 0.807  & 0.738  & 0.814  & 0.805  & 0.913  & 0.924  & 0.925  & 0.923  & 0.927  & 0.939  & \textcolor[rgb]{ 1,  0,  0}{\textbf{0.944}} & 0.930  & 0.936  & 0.936  & \textbf{0.941} & \textcolor[rgb]{ 0,  .69,  .941}{\textbf{0.942}} \\
          \cline{2-18}
          & Rank $\downarrow$ & 15    & 15    & 13    & 13    & 12    & 11    & 10    & 9     & 8     & 4     & \textcolor[rgb]{ 0,  .69,  .941}{\textbf{2}} & 7     & 5     & 6     & \textbf{3} & \textcolor[rgb]{ 1,  0,  0}{\textbf{1}} \\
          \cline{1-18}
    \multirow{5}[0]{*}{DUTLF-D} & $F_\beta^{\text{max}}$ $\uparrow$ &  0.728  & 0.652  & 0.212  & 0.419  & 0.811  & 0.782  & 0.804  & 0.740  & \textcolor[rgb]{ 0,  .69,  .941}{\textbf{0.887}} & 0.748  & 0.884  & 0.882  & 0.836  & 0.869  & \textbf{0.885} & \textcolor[rgb]{ 1,  0,  0}{\textbf{0.912}} \\
          & MAE $\downarrow$   & 0.293  & 0.162  & 0.320  & 0.232  & 0.095  & 0.100  & 0.092  & 0.100  & 0.053  & 0.099  & \textbf{0.053} & 0.054  & 0.064  & 0.054  & \textcolor[rgb]{ 0,  .69,  .941}{\textbf{0.048}} & \textcolor[rgb]{ 1,  0,  0}{\textbf{0.041}} \\
          & $S_\alpha$ $\uparrow$ & 0.659  & 0.568  & 0.361  & 0.499  & 0.831  & 0.801  & 0.808  & 0.749  & 0.888  & 0.775  & \textcolor[rgb]{ 1,  0,  0}{\textbf{0.906}} & \textbf{0.903} & 0.863  & 0.889  & \textcolor[rgb]{ 1,  0,  0}{\textbf{0.906}} & 0.896  \\
          & $E_\xi^{\text{max}}$ $\uparrow$ & 0.800  & 0.734  & 0.590  & 0.654  & 0.899  & 0.856  & 0.861  & 0.811  & 0.933  & 0.834  & \textbf{0.943} & 0.937  & 0.904  & 0.931  & \textcolor[rgb]{ 0,  .69,  .941}{\textbf{0.946}} & \textcolor[rgb]{ 1,  0,  0}{\textbf{0.950}} \\
          \cline{2-18}
          & Rank $\downarrow$ & 13    & 14    & 16    & 15    & 8     & 10    & 9     & 12    & 4     & 11    & \textbf{3} & 5     & 7     & 6     & \textcolor[rgb]{ 0,  .69,  .941}{\textbf{2}} & \textcolor[rgb]{ 1,  0,  0}{\textbf{1}} \\
          \cline{1-18}
    \multirow{5}[0]{*}{NLPR} & $F_\beta^{\text{max}}$ $\uparrow$ &  0.680  & 0.693  & 0.664  & 0.706  & 0.841  & 0.863  & 0.877  & 0.888  & 0.888  & 0.907  & \textcolor[rgb]{ 1,  0,  0}{\textbf{0.925}} & 0.910  & 0.915  & 0.907  & \textcolor[rgb]{ 0,  .69,  .941}{\textbf{0.921}} & \textbf{0.916} \\
          & MAE $\downarrow$   & 0.316  & 0.104  & 0.163  & 0.112  & 0.056  & 0.044  & 0.041  & 0.036  & 0.031  & 0.030  & \textcolor[rgb]{ 1,  0,  0}{\textbf{0.022}} & 0.030  & \textbf{0.025} & 0.031  & \textcolor[rgb]{ 0,  .69,  .941}{\textbf{0.024}} & \textbf{0.025} \\
          & $S_\alpha$ $\uparrow$ & 0.573  & 0.631  & 0.684  & 0.729  & 0.860  & 0.874  & 0.886  & 0.888  & 0.899  & 0.912  & \textcolor[rgb]{ 0,  .69,  .941}{\textbf{0.925}} & 0.915  & \textbf{0.920} & 0.909  & \textcolor[rgb]{ 1,  0,  0}{\textbf{0.927}} & \textbf{0.920} \\
          & $E_\xi^{\text{max}}$ $\uparrow$ & 0.808  & 0.763  & 0.800  & 0.795  & 0.929  & 0.925  & 0.941  & 0.932  & 0.947  & 0.953  & \textcolor[rgb]{ 1,  0,  0}{\textbf{0.963}} & 0.953  & 0.956  & 0.949  & \textcolor[rgb]{ 0,  .69,  .941}{\textbf{0.962}} & \textbf{0.961} \\
          \cline{2-18}
          & Rank $\downarrow$ & 16    & 14    & 15    & 13    & 12    & 11    & 10    & 9     & 8     & 6     & \textcolor[rgb]{ 1,  0,  0}{\textbf{1}} & 5     & 4     & 7     & \textcolor[rgb]{ 0,  .69,  .941}{\textbf{2}} & \textbf{3} \\
          \cline{1-18}
    \multirow{5}[0]{*}{STERE} & $F_\beta^{\text{max}}$ $\uparrow$ &   0.738  & 0.752  & 0.682  & 0.789  & 0.848  & 0.875  & 0.878  & 0.889  & 0.895  & 0.904  & \textcolor[rgb]{ 1,  0,  0}{\textbf{0.913}} & 0.895  & \textcolor[rgb]{ 0,  .69,  .941}{\textbf{0.908}} & 0.895  & \textcolor[rgb]{ 0,  .69,  .941}{\textbf{0.908}} & 0.906  \\
          & MAE $\downarrow$   & 0.301  & 0.172  & 0.197  & 0.148  & 0.086  & 0.064  & 0.060  & 0.051  & 0.047  & 0.046  & \textcolor[rgb]{ 0,  .69,  .941}{\textbf{0.040}} & 0.051  & \textcolor[rgb]{ 1,  0,  0}{\textbf{0.039}} & 0.048  & 0.042  & \textbf{0.041} \\
          & $S_\alpha$ $\uparrow$ & 0.642  & 0.562  & 0.692  & 0.731  & 0.848  & 0.875  & 0.871  & 0.879  & 0.886  & 0.899  & \textcolor[rgb]{ 0,  .69,  .941}{\textbf{0.903}} & 0.890  & \textcolor[rgb]{ 0,  .69,  .941}{\textbf{0.903}} & 0.892  & \textcolor[rgb]{ 1,  0,  0}{\textbf{0.904}} & \textcolor[rgb]{ 0,  .69,  .941}{\textbf{0.903}} \\
          & $E_\xi^{\text{max}}$ $\uparrow$ & 0.811  & 0.771  & 0.806  & 0.819  & 0.912  & 0.925  & 0.923  & 0.925  & 0.938  & 0.938  & \textcolor[rgb]{ 1,  0,  0}{\textbf{0.947}} & 0.932  & \textcolor[rgb]{ 0,  .69,  .941}{\textbf{0.944}} & 0.930  & \textcolor[rgb]{ 0,  .69,  .941}{\textbf{0.944}} & 0.940  \\
          \cline{2-18}
          & Rank $\downarrow$ & 14    & 14    & 14    & 13    & 12    & 10    & 10    & 9     & 6     & 5     & \textcolor[rgb]{ 0,  .69,  .941}{\textbf{2}} & 7     & \textcolor[rgb]{ 1,  0,  0}{\textbf{1}} & 7     & \textbf{3} & 4  \\
          \cline{1-18}
    \multirow{5}[0]{*}{SSD} & $F_\beta^{\text{max}}$ $\uparrow$ &  0.720  & 0.633  & 0.709  & 0.755  & 0.744  & 0.833  & 0.835  & 0.801  & 0.858  & 0.856  & 0.860  & \textcolor[rgb]{ 0,  .69,  .941}{\textbf{0.878}} & \textcolor[rgb]{ 1,  0,  0}{\textbf{0.881}} & \textcolor[rgb]{ 0,  .69,  .941}{\textbf{0.878}} & 0.870  & 0.863  \\
          & MAE $\downarrow$   & 0.313  & 0.195  & 0.204  & 0.169  & 0.098  & 0.062  & 0.063  & 0.082  & 0.059  & 0.059  & 0.053  & 0.053  & \textcolor[rgb]{ 1,  0,  0}{\textbf{0.049}} & \textcolor[rgb]{ 0,  .69,  .941}{\textbf{0.050}} & \textbf{0.052} & \textbf{0.052} \\
          & $S_\alpha$ $\uparrow$ & 0.602  & 0.566  & 0.675  & 0.704  & 0.776  & 0.841  & 0.839  & 0.807  & 0.857  & 0.857  & 0.860  & \textbf{0.868} & 0.866  & \textcolor[rgb]{ 0,  .69,  .941}{\textbf{0.869}} & \textcolor[rgb]{ 1,  0,  0}{\textbf{0.870}} & 0.862  \\
          & $E_\xi^{\text{max}}$ $\uparrow$ & 0.769  & 0.717  & 0.785  & 0.786  & 0.865  & 0.894  & 0.897  & 0.852  & 0.906  & \textcolor[rgb]{ 0,  .69,  .941}{\textbf{0.910}} & 0.902  & \textbf{0.909} & 0.907  & 0.907  & 0.907  & \textcolor[rgb]{ 1,  0,  0}{\textbf{0.914}} \\
          \cline{2-18}
          & Rank $\downarrow$ &
          15    & 16    & 14    & 13    & 12    & 9     & 9     & 11    & 8     & 6     & 7     & \textbf{3} & \textcolor[rgb]{ 1,  0,  0}{\textbf{1}} & \textbf{3} & \textcolor[rgb]{ 0,  .69,  .941}{\textbf{2}} & 5  \\
          \cline{1-18}
    \multirow{5}[0]{*}{SIP} & $F_\beta^{\text{max}}$ $\uparrow$ &  0.720  & 0.634  & 0.788  & 0.680  & 0.717  & 0.860  & 0.849  & 0.869  & 0.852  & 0.880  & \textcolor[rgb]{ 1,  0,  0}{\textbf{0.903}} & 0.891  & 0.896  & \textcolor[rgb]{ 0,  .69,  .941}{\textbf{0.900}} & 0.895  & \textbf{0.898} \\
          & MAE $\downarrow$   & 0.303  & 0.184  & 0.175  & 0.186  & 0.140  & 0.071  & 0.075  & 0.064  & 0.086  & 0.063  & \textcolor[rgb]{ 1,  0,  0}{\textbf{0.049}} & 0.057  & \textcolor[rgb]{ 0,  .69,  .941}{\textbf{0.051}} & 0.054  & \textcolor[rgb]{ 0,  .69,  .941}{\textbf{0.051}} & 0.053  \\
          & $S_\alpha$ $\uparrow$ & 0.616  & 0.511  & 0.732  & 0.683  & 0.716  & 0.842  & 0.835  & 0.850  & 0.806  & 0.860  & \textcolor[rgb]{ 0,  .69,  .941}{\textbf{0.880}} & 0.872  & 0.875  & \textbf{0.878} & \textcolor[rgb]{ 1,  0,  0}{\textbf{0.884}} & 0.873  \\
          & $E_\xi^{\text{max}}$ $\uparrow$ & 0.770  & 0.716  & 0.838  & 0.743  & 0.829  & 0.901  & 0.895  & 0.903  & 0.875  & 0.909  & \textcolor[rgb]{ 0,  .69,  .941}{\textbf{0.925}} & 0.919  & 0.919  & \textbf{0.921} & \textcolor[rgb]{ 1,  0,  0}{\textbf{0.928}} & 0.916  \\
          \cline{2-18}
          & Rank $\downarrow$ & 14    & 16    & 12    & 15    & 13    & 9     & 10    & 8     & 11    & 7     & \textcolor[rgb]{ 1,  0,  0}{\textbf{1}} & 6     & 4     & \textbf{3} & \textcolor[rgb]{ 0,  .69,  .941}{\textbf{2}} & 5  \\
          \Xhline{1pt}
    \end{tabular}}
\end{table*}

\section{Experiments}
We first provide the experimental setup in \S\ref{sec:exp_setup}.
Then, we compare with \sArt RGB-D SOD methods in \S\ref{sec:comparison} and
conduct comprehensive ablation studies in \S\ref{sec:ablation}.
We also discuss the applications of IDR in \S\ref{sec:application}.

\subsection{Experimental Setup}
\label{sec:exp_setup}

\textbf{Implementation details}.
We implement our network in PyTorch~\cite{paszke2019pytorch} and Jittor~\cite{hu2020jittor}.
If not specified, we use MobileNetV2~\cite{sandler2018mobilenetv2} as our backbone.
The $M$ values in the depth stream, CPR, and IDR are set to 4, 4, and 6, respectively.
We resize both RGB and depth images into $320\times 320$.
We use horizontal flipping and random cropping as the default data augmentation for the ablation study.
After freezing the designs and parameters, we apply multi-scale training, \ie, each image is resized into $[256, 288, 320]$ in training, but we keep the size of test images unchanged.
We use a single RTX 2080Ti GPU for training and testing.
The initial learning rate $lr$ is 0.0001, 
and the batch size is 10.
We train our network for 60 epochs.
The $poly$ learning rate policy is applied,
so that the actual learning rate for each epoch $cur\_epoch$ is $(1-\frac{cur\_epoch}{60})^{power}\times lr$, 
where $power$ is 0.9.
The Adam optimizer~\cite{kingma2014adam} is used for optimizing our networks, and the momentum, weight decay, $\beta_1$, and $\beta_2$ are set as 0.9, 0.0001, 0.9, and 0.99, respectively.

\renewcommand{\AddImg}[1]{\includegraphics[width=.06\textwidth]{#1}}
\newcommand{\AddImgs}[1]{\AddImg{#1} & \AddImg{#1_depth} & \AddImg{#1_gt} &  
   \AddImg{#1_ours} & \AddImg{#1_cpfp} & \AddImg{#1_jldcf} & \AddImg{#1_s2ma} & 
   \AddImg{#1_ucnet}
}

\CheckRmv{
\begin{figure}[t!]
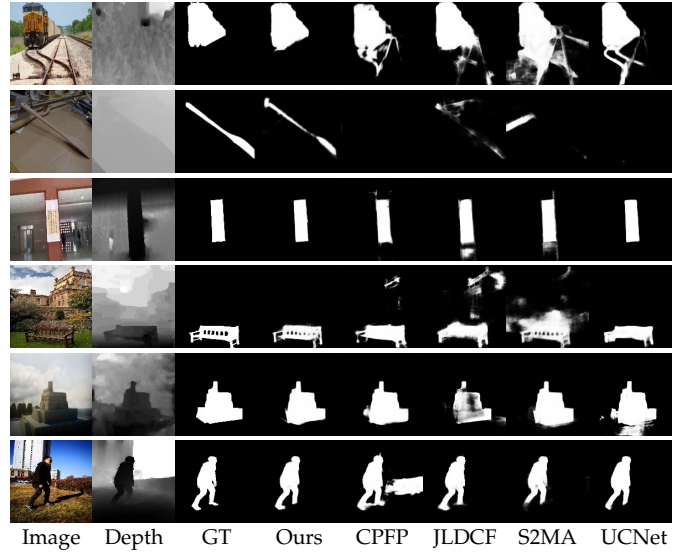

  \centering
  \renewcommand{\arraystretch}{0.6}
  \setlength{\tabcolsep}{0.04mm}
  \begin{tabular}{cccccccc}
    \AddImgs{NJU2K/000417_left} \\
    \AddImgs{DUT/0719} \\
    \AddImgs{NLPR/1_02-54-18} \\
    \AddImgs{STERE/image_left78} \\
    \AddImgs{SSD/histrory10301} \\
    \AddImgs{SIP/48} \\
    \footnotesize Image & \footnotesize Depth &\footnotesize  GT
     &\footnotesize Ours &\footnotesize CPFP &\footnotesize JLDCF 
     &\footnotesize S2MA &\footnotesize UCNet \\
  \end{tabular}
  \vspace{-8pt}
  \caption{\textbf{Qualitative comparison of six challenging datasets.}
    The results from top to bottom are from 
    NJU2K, DUTLF-D, NLPR, STERE, SSD, and SIP datasets, respectively.
}\label{fig:visualcomp}
\end{figure}
}

\CheckRmv{
\begin{table}[t]
  \centering
  \caption{\textbf{CPU inference time of different methods}. Default input size of each method is applied to test CPU inference time.}
  \label{tab:cpu_runtime}%
  \setlength\tabcolsep{6pt}
  \renewcommand\arraystretch{1.0}
\resizebox{.47\textwidth}{!}{
    \begin{tabular}{cccc}
      \Xhline{3\arrayrulewidth}    Method   & Ours     & JLDCF~\cite{fu2020jl}   & UCNet~\cite{zhang2020uc} \\ 
    Input Size & $320\times 320$ & $320\times 320$ & $352\times 352$ \\
    Inference Time (ms) & \textbf{43} (1$\times$)  & 7246 (150$\times$)   & 784 (18$\times$)   \\
    \Xhline{3\arrayrulewidth}   Method  & D3Net~\cite{fan2020rethinking}   & S2MA~\cite{liu2020learning}    & DMRA~\cite{piao2019depth}
   \\
   Input Size & $224\times 224$ & $256\times 256$ & $256\times 256$ \\
   Inference Time (ms) & 677 (15$\times$) & 3049 (70$\times$)    & 2381 (55$\times$)    \\
  \Xhline{3\arrayrulewidth}    \end{tabular}%
}
\end{table}%
}

\CheckRmv{
\begin{table}[!t]
  \centering
  \caption{
    \textbf{Quantitative comparisons of restored depth maps by IDR
 with different scales of input depth maps}.
  The ground-truth depth (GT) recovered from different resolutions using different interpolation methods are compared.}
  \label{tab:depth_quality}
  \setlength\tabcolsep{10pt}
\resizebox{.47\textwidth}{!}{
    \begin{tabular}{ccccccc}
      \Xhline{3\arrayrulewidth}
Setting & (a) & (b)  &  (c)  & (d)  \\
 Scale & 1/16 & 1/8 & 1/32 & 1/8\\
    Data & IDR  &    GT &  GT     &  GT    \\
     Up Type   & Bilinear    & Bilinear     & Bilinear  & Nearest  \\
     \Xhline{3\arrayrulewidth}
    PSNR & 22.86 &  30.17  & 22.55  &  24.27 \\
         
    SSIM   & .8687 & .9194 & .8445 &  .8170 \\
    \Xhline{3\arrayrulewidth}
    \end{tabular}}
\end{table}%
}

\textbf{Datasets}.
We conduct experiments on seven widely-used datasets, including NJU2K~\cite{ju2014depth}, DUTLF-D~\cite{piao2019depth}, 
NLPR~\cite{peng2014rgbd}, STERE~\cite{niu2012leveraging}, 
 SSD~\cite{zhu2017three}, 
and SIP~\cite{fan2020rethinking}.
They contain 1985, 1200, 1000, 1000, 80, 927 images, respectively.
Following
\cite{zhao2019constrast, liu2020learning, zhang2020uc, zhao2020single}, 
we use 1500 images of NJU2K~\cite{ju2014depth} 
and 700 images of NLPR~\cite{peng2014rgbd} for training,
and the other 485 images of NJU2K~\cite{ju2014depth} 
and 300 images of NLPR ~\cite{peng2014rgbd} for testing.
Except for DUTLF-D~\cite{piao2019depth}, other datasets are directly used for testing.
On the DUTLF-D~\cite{piao2019depth} dataset, we follow~\cite{piao2019depth, zhao2020single} to use 800 images for training and the other 400 images for testing.

\textbf{Evaluation metrics}.
Following recent works
\cite{chen2018progressively, chen2019three, zhao2019constrast, zhang2020uc},
we adopt two widely-used metrics for evaluation. 
The first is F-measure $F_\beta$, where
$\beta$ is set as 0.3 to emphasize the significance of the precision,
as suggested by
\cite{achanta2009frequency,zhao2019constrast,piao2019depth, zhao2020single}. 
We compute the maximum $F_\beta$ as $F_\beta^{\text{max}}$ under different thresholds.
Higher $F_\beta^{\text{max}}$ indicates better performance.
The second is the Mean Absolute Error (MAE), which is the lower, the better.
We also report scores of recently proposed S-measure  $S_\alpha$ \cite{fan2017structure} 
and maximum E-measure $E_\xi^{\text{max}}$ \cite{fan2018enhanced} under different thresholds for reference.
We follow the official paper to compute $S_\alpha$ and $E_\xi^{\text{max}}$.
We calculate the overall rank of each method on each dataset based on the above four metrics.
Besides,
we report the number of parameters and running time of each method for efficiency analysis.

\CheckRmv{
\begin{table}[!t]
  \centering
  \caption{\textbf{Ablation study for the RGB-D fusion and IDR branch}. 
   Note that the variants of No. 6 and 12 (in \textbf{bold}) fuse RGB and depth features only at the coarsest feature level.}
  \label{tab:rgbd-fusion}
  \setlength\tabcolsep{7pt}
  \resizebox{.47\textwidth}{!}{
    \begin{tabular}{rrrrrrrrr}
      \Xhline{1pt}
    \multicolumn{1}{c}{\multirow{2}[0]{*}{No.}} & \multicolumn{5}{c}{Features to be fused} & \multicolumn{1}{c}{\multirow{2}[0]{*}{IDR}} 
    &\multicolumn{1}{c}{\multirow{2}[0]{*}{$F_\beta^{\text{max}}$}} & \multicolumn{1}{c}{\multirow{2}[0]{*}{MAE}}\\
    & $\mathcal{C}_1$     & $\mathcal{C}_2$    & $\mathcal{C}_3$    & $\mathcal{C}_4$     & $\mathcal{C}_5$   &       &       &  \\
          \Xhline{1pt}
    1     & \ding{52} & \ding{52} & \ding{52} & \ding{52} & \ding{52} & \ding{52} & 0.899  & 0.052  \\
    2     & \ding{52} & \ding{52} &       &       &       & \ding{52} & 0.894  & 0.050  \\
    3     & \ding{52} & \ding{52} & \ding{52} &       &       & \ding{52} & 0.897  & 0.047  \\
    4     &       & \ding{52} & \ding{52} & \ding{52} &       & \ding{52} & 0.902  & 0.048  \\
    5     &       &       & \ding{52} & \ding{52} & \ding{52} & \ding{52} & 0.902  & 0.046  \\
    6     &       &       &       &       & \ding{52} & \ding{52} & \textbf{0.906} & \textbf{0.045} \\
    7     & \ding{52} & \ding{52} & \ding{52} & \ding{52} & \ding{52} &       & 0.895  & 0.047  \\
    8     & \ding{52} & \ding{52} &       &       &       &       & 0.892  & 0.049  \\
    9     & \ding{52} & \ding{52} & \ding{52} &       &       &       & 0.896  & 0.048  \\
    10    &       & \ding{52} & \ding{52} & \ding{52} &       &       & 0.895  & 0.048  \\
    11    &       &       & \ding{52} & \ding{52} & \ding{52} &       & 0.898  & 0.048  \\
    12    &       &       &       &       & \ding{52} &       & \textbf{0.896} & \textbf{0.047} \\
    13    &       &       &       &       &       &       & 0.887  & 0.052  \\
    \Xhline{1pt}
    \end{tabular}}
\end{table}%
}

\begin{table}[!t]
  \centering
  \caption{\textbf{Comparison of RGB-D fusion strategies}. The results of default fusion strategy are with \textbf{bold} fonts.}
  \label{tab:single_fusion_comp}
  \setlength\tabcolsep{12pt}
  \resizebox{.47\textwidth}{!}{
    \begin{tabular}{lrrrr}
      \Xhline{1pt}
    \multicolumn{1}{c}{\multirow{2}[0]{*}{Metric}} & \multicolumn{2}{c}{Single Stream} & \multicolumn{2}{c}{Two Streams} \\
    &  IDR \ding{52} &  IDR \ding{54}     & IDR \ding{52} &  IDR \ding{54} \\
    \Xhline{1pt}
    $F_\beta^{\text{max}}$ & 0.900  & 0.894  & \textbf{0.906} & \textbf{0.896} \\
    MAE   & 0.048  & 0.051  & \textbf{0.045} & \textbf{0.047} \\
    \Xhline{1pt}
    \end{tabular}}
\end{table}

\subsection{Comparison to \sArt methods}\label{sec:comparison}

We first compare our method with 15 recent \sArt methods published on six widely-used datasets. 
Most methods are based on VGG-16 \cite{simonyan2014very}, 
except the ResNet-101-based JLDCF and ResNet-50-based UCNet.
The saliency maps of other methods are from their released results if provided,
otherwise they are computed by their released models.

\textbf{Quantitative comparison.}
\tabref{tab:benchmark} shows the results.
Our method runs at 450fps and only has 6.5M parameters.
Other methods are much slower and heavier than our method.
For example, JLDCF~\cite{fu2020jl} is 50$\times$ slower and has 20$\times$ more parameters. 
UCNet is 26$\times$ slower and has 5.5$\times$ more parameters.
Besides,
our method outperforms other methods on 
NJU2K~\cite{ju2014depth} and DUTLF-D~\cite{piao2019depth} datasets,
and ranks from 3$^{\text{rd}}$ to 5$^{\text{th}}$ on the other 4 datasets.
The above results demonstrate the high efficiency and accuracy of our method.

\renewcommand{\AddImg}[1]{\includegraphics[width=.095\textwidth]{#1}}

\CheckRmv{
\begin{figure}[t]
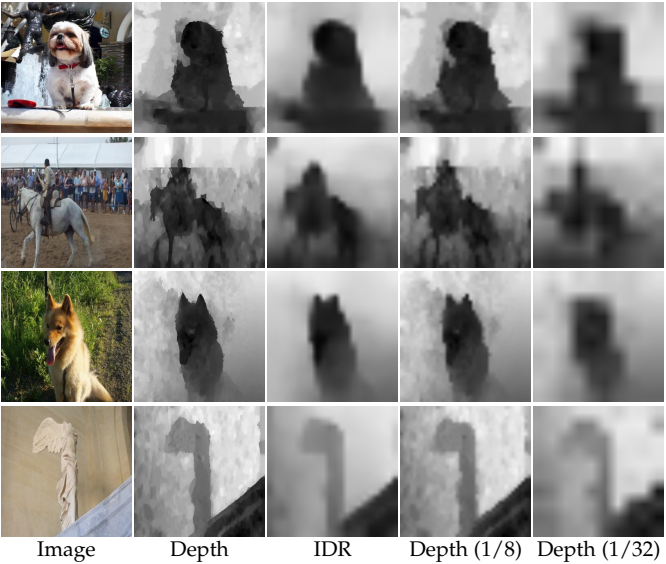

\centering
\renewcommand{\arraystretch}{0.5}
  \setlength{\tabcolsep}{0.2mm}
  \begin{tabular}{ccccccc}
\AddImg{NJU2K/000205_left}   & \AddImg{NJU2K/000205_left_depth}
& \AddImg{NJU2K/000205_left_depth_pred}   & \AddImg{NJU2K/000205_left_depth_8bl_gt}
& \AddImg{NJU2K/000205_left_depth_32bl_gt}   
\\
\AddImg{NJU2K/000424_left}   & \AddImg{NJU2K/000424_left_depth} & \AddImg{NJU2K/000424_left_depth_pred}
  & \AddImg{NJU2K/000424_left_depth_8bl_gt} 
  & \AddImg{NJU2K/000424_left_depth_32bl_gt}  
\\
\AddImg{NJU2K/000515_left}   & \AddImg{NJU2K/000515_left_depth} 
& \AddImg{NJU2K/000515_left_depth_pred}  & \AddImg{NJU2K/000515_left_depth_8bl_gt}  
& \AddImg{NJU2K/000515_left_depth_32bl_gt} 
\\
\AddImg{NJU2K/001046_left}   & \AddImg{NJU2K/001046_left_depth} & \AddImg{NJU2K/001046_left_depth_pred}  & \AddImg{NJU2K/001046_left_depth_8bl_gt}  & \AddImg{NJU2K/001046_left_depth_32bl_gt}  
\\
\footnotesize Image &\footnotesize Depth 
&\footnotesize IDR &\footnotesize Depth (1/8) & \footnotesize Depth (1/32)
\end{tabular}
\caption{\textbf{Visual comparisons of restored depth maps by IDR
 with different scales of input depth maps.}
Results of the last 3 columns have been upsampled with bilinear interpolation to match the size of the input depth map.}
\label{fig:depth_visual_cmp}
\end{figure}
}

\textbf{Qualitative comparison.}
\figref{fig:visualcomp} shows the results.
Due to the limited space, here we only compare our method with 
CPFP~\cite{zhao2019constrast}, JLDCF~\cite{fu2020jl}, 
S2MA~\cite{liu2020learning}, and UCNet~\cite{zhang2020uc} on all involved datasets.
Our method can work well on several kinds of complex scenarios with noisy depth information, 
while others may fail in such scenarios.

\textbf{CPU inference time.}
We test the inference time for different methods in a single core of Intel i7 8700K@3.7GHz CPU.
The results are shown in \tabref{tab:cpu_runtime}.
While the inference time of other \sArt methods (677 $\sim$ 7246 ms) is far from the bar of real-time speed ($\sim$50 ms), 
the CPU inference time of our method can achieve a real-time speed of 43 ms for each RGB-D input.

\subsection{Ablation Study}\label{sec:ablation}

We evaluate each proposed component on the test split of the NJU2K~\cite{ju2014depth} dataset.
$F_\beta^{\text{max}}$ and MAE are applied as the primary metrics.
The results are analyzed below.

\textbf{Different RGB-D fusions}. 
\tabref{tab:rgbd-fusion} shows the results for the RGB-D fusion at different stages.
When trained with the IDR branch, 
fusing RGB and depth features at the coarsest level 
results in the best performance (No. 6).
Fusing RGB and depth features in the last three levels results in the best performance when trained without the IDR branch (No. 11). 
The IDR branch substantially improves the performance in most cases (No. 3 - 6), 
and only MAE (fusing low-level features) degrades (No. 1, 2).
This validates the efficacy of the combination of our proposed fusion strategy and the IDR branch.
We also compare the used fusion strategy 
with the early fusion strategy, 
which concatenates the input RGB image and depth map at the input stage.
Although the latter strategy is more efficient, 
our initial fusion strategy significantly outperforms it 
(\tabref{tab:single_fusion_comp}). 
Hence,
To ensure accuracy and efficiency, we fuse RGB and depth features at the coarsest level. 

\textbf{Saved time for RGB-D fusion.}
In \methodname, we conduct RGB and depth information fusion only at the coarsest level.
Conducting RGB-D feature fusion of all levels  will result in 260fps for our method. Hence, we will save 42\% ($1-\frac{260}{450}$) time. 
In other words, fusing only the coarsest will speed up our method by 73\% ($\frac{450}{260} - 1$).

\textbf{Depth restoration quality}.
We explore the restored depth quality of IDR with the widely-used PSNR and SSIM~\cite{wang2004image} metrics.
The scale of the restored depth map in IDR is 1/16 of the input depth map.
For comparison, we evaluate the quality of the nearest and bilinear interpolation of the input depth map of 1/8 scale.
As the IDR branch receives depth features of the 1/32 scale, 
we also report the quality of the bilinear interpolation of depth GT with the 1/32 scale.
Results are shown in \tabref{tab:depth_quality}.
We observe that the restored depth map is closer
to the input depth map than the upsampled 1/32 GT.
We also conduct visual comparisons of the above settings in \figref{fig:depth_visual_cmp}.
One can see that the restored depth maps from IDR
keep good restoration quality and are with less noise than the upsampled 1/8 GT.

\textbf{Loss selection in IDR.}
As introduced in \secref{sec:IDR}, instead of using the trivial L1/L2 loss, the SSIM metric \cite{wang2004image} 
is chosen as the loss of IDR. 
We validate this design via training our MobileSal with L1/L2/SSIM loss.
We find that IDR with the L1/L2 loss can enhance 
MobileSal by 0.5\%/0.7\% in terms of the $F_\beta^{\text{max}}$, 
and is 0.5\%/0.3\% lower than IDR with the SSIM loss.
This is because SSIM loss provides the structural similarity rather than the simple point-to-point error computed by the  L1/L2 loss.
Based on the above discussion, we choose the SSIM loss as the supervision of IDR.

\textbf{The $\lambda$ coefficient in the loss function}.
$\lambda$ decides the loss weight of the depth restoration loss, as described in \equref{equ:loss_combine}. 
We conduct experiments on our method with different $\lambda$ settings.
The results are shown in \tabref{tab:loss_lambda}.
The IDR branch brings substantial improvement to the robustness of our method with different $\lambda$. 
Since the third column achieves the best performance,
we adopt $\lambda=0.3$ as the default setting for training our network.

\begin{table}[!t]
  \centering
  \caption{\textbf{Ablation study for $\lambda$ coefficient selection}. 
}
   \label{tab:loss_lambda}
  \setlength\tabcolsep{10pt}
\resizebox{\columnwidth}{!}{
    \begin{tabular}{ccccccc}
      \Xhline{1pt}
    No.   & 1     & 2     & 3     & 4     & 5     & 6  \\
    $\lambda$ & 0     & 0.1   & 0.3   & 0.5   & 1.0   & 2.0  \\
    \Xhline{1pt}
    $F_\beta^{\text{max}}$ & 0.896  & 0.902  & \textbf{0.906} & 0.902  & 0.903  & 0.902  \\
    MAE   & 0.047  & 0.046  & \textbf{0.045} & 0.046  & 0.044  & 0.046  \\
    \Xhline{1pt}
    \end{tabular}}
\end{table}%

\begin{table}[!t]
  \centering
  \caption{\textbf{Efficacy of CMF}. ``RGB'' denotes the RGB backbone with the RGB input. ``Depth'' indicates the network with the depth stream and depth input.}
  \label{tab:cmf_module}
\setlength\tabcolsep{8pt}
\resizebox{.47\textwidth}{!}{
    \begin{tabular}{cccccrr}
      \Xhline{1pt}
    \multicolumn{1}{l}{No.} & RGB & CPR   & Depth     & CMF    & $F_\beta^{\text{max}}$ &  MAE \\
    \Xhline{1pt}
    1     & \ding{52} &       &       &       & 0.852  & 0.068  \\
    2     & \ding{52} & \ding{52} &       &       & 0.887  & 0.052  \\
    3     & \ding{52} & \ding{52} & \ding{52} &       & 0.894  & 0.048  \\
    4     & \ding{52} & \ding{52} & \ding{52} & \ding{52} & \textbf{0.906}  & \textbf{0.045}  \\
    \Xhline{1pt}
    \end{tabular}}
\end{table}

\begin{table}[!t]
  \centering
  \caption{\textbf{Comparison of different operations for the initial RGB-D fusion in \equref{equ:initial_rgbd_fuse}. 
  ``Multiplication'' and ``Addition'' operations are element-wise. 
  Concatenation operation is along channels.
  The results of the default fusion strategy are with \textbf{bold} fonts.}}
  \label{tab:initial_rgbd_fuse}
  \setlength\tabcolsep{5pt}
  \resizebox{.48\textwidth}{!}{
    \begin{tabular}{lrrrrrr}
      \Xhline{1pt}
    Operation & \multicolumn{2}{c}{Multiplication} & \multicolumn{2}{c}{Addition} & \multicolumn{2}{c}{Concatenation} \\
    Metric &  $F_\beta^{\text{max}}$ &  MAE      &  $F_\beta^{\text{max}}$ &  MAE &  $F_\beta^{\text{max}}$ &  MAE \\
    \Xhline{1pt}
    Results & \textbf{0.906}  & \textbf{0.045}  & 0.897 & 0.048 & 0.900 & 0.046 \\
    \Xhline{1pt}
    \end{tabular}}
\end{table}

\textbf{Depth information and the CMF module}.
The results in \tabref{tab:cmf_module} demonstrate the effects of depth information and the CMF module.
Results with depth input are trained with the IDR branch.
The providence of depth maps without the CMF module 
only uses an element-wise multiplication for RGB-D fusion.
The results show that the depth information is very helpful for RGB-D SOD even with a very simple operation.
More specifically, we also observe substantial improvement with the CMF module.

\textbf{Operation for initial RGB-D fusion in the CMF module.}
As formulated in \equref{equ:initial_rgbd_fuse}, an element-wise multiplication is applied to the initial RGB-D fusion.
To validate the effectiveness of the selected operation,
we compare this design with the widely-used element-wise addition.
The results are presented in \tabref{tab:initial_rgbd_fuse}.
Element-wise multiplication largely outperforms element-wise addition and concatenation by 0.9\% and 0.6\% in terms of the maximum F-measure, suggesting the superiority of element-wise multiplication in the initial RGB-D fusion of the CMF module.

\textbf{Final fusion strategy of the CMF module}.
As described in \equref{equ:CMF}, both the RGB attention vector $\mathbf{v}$ and RGB-D feature $\mathcal{T}$ join the fusion with depth feature $D_5$.
We validate the effectiveness of the above design via removing $\mathbf{v}$ or $\mathcal{T}$.
Experimental results are shown in \tabref{tab:cmf_fusion}.
We observe a large performance degradation after removing $\mathcal{T}$.
This is because in the feature fusion of \equref{equ:CMF}, only $\mathcal{T}$
contains spatial-wise RGB features.
Besides, $\mathbf{v}$ is an attention vector obtained by RGB features, and $D_5$ is the pure depth feature.
Removing $\mathbf{v}$ also affects the performance a bit because it provides the channel-wise RGB attention in the RGB-D fusion.

\begin{table}[t]
  \centering
  \caption{\textbf{Analysis of the fusion of CMF in \equref{equ:CMF}}. No. 1: trivial RGB-D fusion via element-wise multiplication; No. 2-3: variants of removing $\mathbf{v}$ or $\mathcal{T}$ in \equref{equ:CMF}; No. 4: the default design.}
  \label{tab:cmf_fusion}
\setlength\tabcolsep{9pt}
\resizebox{.47\textwidth}{!}{
    \begin{tabular}{cccccrr}
      \Xhline{1pt}
    \multicolumn{1}{l}{No.} & Depth & CMF & $\mathbf{v}$   & $\mathcal{T}$     & $F_\beta^{\text{max}}$ &  MAE \\
    \Xhline{1pt}
    1     & \ding{52} &  &     &       & 0.894  & 0.048  \\
    2     & \ding{52} &  \ding{52}  & \ding{52}      &       & 0.895  & 0.049  \\
    3     & \ding{52}&   \ding{52} &  & \ding{52}       & 0.902  & 0.046  \\
    4     & \ding{52} &  \ding{52} & \ding{52} & \ding{52}  & \textbf{0.906}  & \textbf{0.045}  \\
    \Xhline{1pt}
    \end{tabular}}
\end{table}

\CheckRmv{
\begin{table}[!t]
  \centering
  \caption{\textbf{Ablation study for the dilation rates of CPR}. ``D Rates'' means the dilation rates of each depthwise separable convolution.}
  \label{tab:cpr}
\setlength\tabcolsep{7pt}
\resizebox{.47\textwidth}{!}{
    \begin{tabular}{ccccccc}\Xhline{1pt}
    No.   & 1     & 2     & 3     & 4     & 5     & 6 \\
    D Rates & 1, 2, 3 & 1     & 2     & 3     & 1, 3, 6 & 1, 4, 8 \\
    \Xhline{1pt}
    $F_\beta^{\text{max}}$ & \textbf{0.906}  & 0.900  & 0.892  & 0.897  & 0.903  & 0.901  \\
    MAE   & \textbf{0.045}  & 0.047  & 0.048  & 0.047  & 0.046  & 0.048  \\
    \Xhline{1pt}
    \end{tabular}}
\end{table}%
}

\textbf{Compact pyramid refinement.}
\tabref{tab:cpr} shows the results for CPR, where different dilation strategies are used.
We test the default setting (No. 1), single convolution with different dilation rates (No. 2 - 4), and convolutions with sparse combinations of dilation rates (No. 5, 6).
The default setting with compact dilation rates (1, 2, 3) significantly outperforms other settings, demonstrating the efficacy of CPR.

\textbf{Hybrid loss.}
To validate the effectiveness of the Dice loss, we test the performance only trained with the binary cross-entropy loss.
We find that adding Dice loss supervision will improve the MAE performance by 0.1\% $\sim$ 0.2\% via providing high contrast, but will not affect the performance of $F_\beta^{\text{max}}$.

\subsection{Application of IDR to other tasks}\label{sec:application}

Our proposed IDR freely strengthens the feature representations of the backbone network, given the RGB-D input in the inference stage.
It is unbound to RGB-D SOD
that is a segmentation task aiming at deriving a saliency map.
To show the potential of IDR on other tasks, we evaluate the performance gain of IDR 
in the RGB-D semantic segmentation.
The goal of this task, \ie, assigning each pixel with semantic labels, is similar to RGB-D SOD that predicts saliency probability of each pixel.

\indent
\textbf{Experimental setup.} 
We select two recent \sArt representatives \cite{chen2021spatial,chen2020bi} as the baselines.
We use the official code provided by the authors to implement our ideas.
Following \cite{chen2020bi,chen2021spatial}, we conduct our experiments on the NYUDv2 dataset \cite{nyudv2}, 
which consists of 1449 RGB images with corresponding depth maps and pixel-level semantic labels containing 40 semantic classes.
This dataset has 795 and 654 images for training and testing, respectively.
The training and testing settings follow the official papers \cite{chen2020bi,chen2021spatial}.
Similar to MobileSal, for \cite{chen2020bi}, RGB features re-calibrated by the depth features for each stage are fed into the IDR branch.
However, for \cite{chen2021spatial}, different from MobileSal,
the output features of the first four stages
are the input of the IDR branch because the features of the last stage in \cite{chen2021spatial} are directly used for the prediction of semantic segmentation.

\indent
\textbf{Evaluation metrics.}
Following \cite{chen2020bi,chen2021spatial},
mean IoU (mIoU), is used as the primary evaluation metric.
We also report the results of 
pixel accuracy (Acc) and mean accuracy (mAcc) for reference. 
Please refer to \cite{chen2021spatial} for more details about 
the computation of the above three metrics.

\indent
\textbf{Experimental results.}
We show the results in \tabref{tab:app_rgbd_semanticseg}.
With the computationally-free IDR incorporated, 
we observe that the performance of both methods \cite{chen2020bi,chen2021spatial} has a large improvement in terms of the mIoU metric.
This suggests that the idea of IDR is also applicable and powerful for RGB-D semantic segmentation without any extra inference cost.

\CheckRmv{
\begin{table}[!t]
  \centering
  \caption{\revise{\textbf{Effect of IDR for the methods of RGB-D semantic segmentation}. 
  Higher values of all metrics indicate better performance.}}
  \label{tab:app_rgbd_semanticseg}
\setlength\tabcolsep{4pt}
\resizebox{.47\textwidth}{!}{
  \begin{tabular}{l|lcc|lcc}
    \Xhline{1pt}
  Method & \multicolumn{3}{c|}{SGNet \cite{chen2021spatial} } & \multicolumn{3}{c}{Chen \etal \cite{chen2020bi}} \\
  Metrics (\%) & mIoU   &  Acc   & mAcc & mIoU   &  Acc  & mAcc   \\
  \Xhline{1pt}
  Baseline & 49.6  & 75.6  & 61.9 & 51.4 & 77.1 & 62.9 \\
  + IDR   & \textbf{50.5(+0.9)}  & \textbf{76.3}  & \textbf{62.2} & \textbf{52.2(+0.8)} & \textbf{77.3} & \textbf{64.0}\\
  \Xhline{1pt}
  \end{tabular}}
\end{table}%
}

\section{Conclusion} 

We propose a new method, \methodname, which aims at efficient RGB-D SOD.
Unlike other accurate RGB-D SOD methods, 
we are the first to shed light upon
efficient RGB-D SOD by proposing an 
extremely efficient network \methodname~with a speed of 450fps.
With less powerful features provided by the mobile backbone network, 
we propose the implicit depth restoration (IDR) technique to 
strengthen the less powerful features learned by mobile backbone networks.
We perform ablation study for the proposed techniques in \methodname~and
demonstrate their effectiveness.
We conduct experimental comparisons with \sArt methods on six popular benchmarks.
The results show that \methodname~performs favorably against \sArt methods,
with fewer parameters and much faster speed.
In terms of CPU inference time, our method is 15$\sim$150$\times$ faster.
Our method can serve as a strong baseline for future efficient RGB-D SOD research.
For future research, we plan to extend our work via pyramid pooling transformer \cite{wu2021p2t} for better performance.

\section{Acknowledgment}

This project is supported by the National Key Research and Development 
Program of China (Grant No. 2018AAA0100400) and NSFC (NO. 61922046).

{
\bibliographystyle{IEEEtran}
\bibliography{reference}

\begin{thebibliography}{10}
\providecommand{\url}[1]{#1}
\csname url@samestyle\endcsname
\providecommand{\newblock}{\relax}
\providecommand{\bibinfo}[2]{#2}
\providecommand{\BIBentrySTDinterwordspacing}{\spaceskip=0pt\relax}
\providecommand{\BIBentryALTinterwordstretchfactor}{4}
\providecommand{\BIBentryALTinterwordspacing}{\spaceskip=\fontdimen2\font plus
\BIBentryALTinterwordstretchfactor\fontdimen3\font minus
  \fontdimen4\font\relax}
\providecommand{\BIBforeignlanguage}[2]{{%
\expandafter\ifx\csname l@#1\endcsname\relax
\typeout{** WARNING: IEEEtran.bst: No hyphenation pattern has been}%
\typeout{** loaded for the language `#1'. Using the pattern for}%
\typeout{** the default language instead.}%
\else
\language=\csname l@#1\endcsname
\fi
#2}}
\providecommand{\BIBdecl}{\relax}
\BIBdecl

\bibitem{hong2015online}
S.~Hong, T.~You, S.~Kwak, and B.~Han, ``Online tracking by learning
  discriminative saliency map with convolutional neural network,'' in
  \emph{Int. Conf. Mach. Learn.}, 2015, pp. 597--606.

\bibitem{wang2018deep}
W.~Wang, J.~Shen, and H.~Ling, ``A deep network solution for attention and
  aesthetics aware photo cropping,'' \emph{IEEE Trans. Pattern Anal. Mach.
  Intell.}, vol.~41, no.~7, pp. 1531--1544, 2018.

\bibitem{liu2020leveraging}
Y.~Liu, Y.-H. Wu, P.-S. Wen, Y.-J. Shi, Y.~Qiu, and M.-M. Cheng, ``Leveraging
  instance-, image-and dataset-level information for weakly supervised instance
  segmentation,'' \emph{IEEE Trans. Pattern Anal. Mach. Intell.}, 2020.

\bibitem{zhang2017amulet}
P.~Zhang, D.~Wang, H.~Lu, H.~Wang, and X.~Ruan, ``Amulet: Aggregating
  multi-level convolutional features for salient object detection,'' in
  \emph{Int. Conf. Comput. Vis.}, 2017, pp. 202--211.

\bibitem{pang2020multi}
Y.~Pang, X.~Zhao, L.~Zhang, and H.~Lu, ``Multi-scale interactive network for
  salient object detection,'' in \emph{IEEE Conf. Comput. Vis. Pattern Recog.},
  2020, pp. 9413--9422.

\bibitem{zhao2020suppress}
X.~Zhao, Y.~Pang, L.~Zhang, H.~Lu, and L.~Zhang, ``Suppress and balance: A
  simple gated network for salient object detection,'' in \emph{Eur. Conf.
  Comput. Vis.}, 2020, pp. 35--51.

\bibitem{chen2019three}
H.~Chen and Y.~Li, ``Three-stream attention-aware network for {RGB-D} salient
  object detection,'' \emph{IEEE Trans. Image Process.}, vol.~28, no.~6, pp.
  2825--2835, 2019.

\bibitem{zhao2019constrast}
J.~Zhao, Y.~Cao, D.-P. Fan, X.-Y. Li, L.~Zhang, and M.-M. Cheng, ``Contrast
  prior and fluid pyramid integration for {RGBD} salient object detection,'' in
  \emph{IEEE Conf. Comput. Vis. Pattern Recog.}, 2019, pp. 3927--3936.

\bibitem{piao2019depth}
Y.~Piao, W.~Ji, J.~Li, M.~Zhang, and H.~Lu, ``Depth-induced multi-scale
  recurrent attention network for saliency detection,'' in \emph{Int. Conf.
  Comput. Vis.}, 2019, pp. 7254--7263.

\bibitem{fan2020rethinking}
D.-P. Fan, Z.~Lin, Z.~Zhang, M.~Zhu, and M.-M. Cheng, ``Rethinking rgb-d
  salient object detection: Models, data sets, and large-scale benchmarks,''
  \emph{IEEE Trans. Neur. Net. Learn. Syst.}, vol.~32, no.~5, pp. 2075--2089,
  2020.

\bibitem{zhang2020uc}
J.~Zhang, D.-P. Fan, Y.~Dai, S.~Anwar, F.~Saleh, S.~Aliakbarian, and N.~Barnes,
  ``Uncertainty inspired {RGB-D} saliency detection,'' \emph{IEEE Trans.
  Pattern Anal. Mach. Intell.}, 2021.

\bibitem{zhao2020single}
X.~Zhao, L.~Zhang, Y.~Pang, H.~Lu, and L.~Zhang, ``A single stream network for
  robust and real-time {RGB-D} salient object detection,'' in \emph{Eur. Conf.
  Comput. Vis.}, 2020, pp. 646--662.

\bibitem{li2020rgb}
C.~Li, R.~Cong, Y.~Piao, Q.~Xu, and C.~C. Loy, ``{RGB-D} salient object
  detection with cross-modality modulation and selection,'' in \emph{Eur. Conf.
  Comput. Vis.}, 2020, pp. 225--241.

\bibitem{fu2020jl}
K.~Fu, D.-P. Fan, G.-P. Ji, Q.~Zhao, J.~Shen, and C.~Zhu, ``Siamese network for
  {RGB-D} salient object detection and beyond,'' \emph{IEEE Trans. Pattern
  Anal. Mach. Intell.}, 2021.

\bibitem{ju2014depth}
R.~Ju, L.~Ge, W.~Geng, T.~Ren, and G.~Wu, ``Depth saliency based on anisotropic
  center-surround difference,'' in \emph{Int. Conf. Image Process.}, 2014, pp.
  1115--1119.

\bibitem{howard2017mobilenets}
A.~G. Howard, M.~Zhu, B.~Chen, D.~Kalenichenko, W.~Wang, T.~Weyand,
  M.~Andreetto, and H.~Adam, ``Mobile{N}ets: Efficient convolutional neural
  networks for mobile vision applications,'' \emph{arXiv preprint
  arXiv:1704.04861}, 2017.

\bibitem{sandler2018mobilenetv2}
M.~Sandler, A.~Howard, M.~Zhu, A.~Zhmoginov, and L.-C. Chen, ``Mobile{N}et{V}2:
  Inverted residuals and linear bottlenecks,'' in \emph{IEEE Conf. Comput. Vis.
  Pattern Recog.}, 2018, pp. 4510--4520.

\bibitem{zhang2018shufflenet}
X.~Zhang, X.~Zhou, M.~Lin, and J.~Sun, ``Shuffle{N}et: An extremely efficient
  convolutional neural network for mobile devices,'' in \emph{IEEE Conf.
  Comput. Vis. Pattern Recog.}, 2018, pp. 6848--6856.

\bibitem{ma2018shufflenet}
N.~Ma, X.~Zhang, H.-T. Zheng, and J.~Sun, ``Shuffle{N}et v2: Practical
  guidelines for efficient cnn architecture design,'' in \emph{Eur. Conf.
  Comput. Vis.}, 2018, pp. 116--131.

\bibitem{simonyan2014very}
K.~Simonyan and A.~Zisserman, ``Very deep convolutional networks for
  large-scale image recognition,'' in \emph{Int. Conf. Learn. Represent.},
  2015.

\bibitem{he2016deep}
K.~He, X.~Zhang, S.~Ren, and J.~Sun, ``Deep residual learning for image
  recognition,'' in \emph{IEEE Conf. Comput. Vis. Pattern Recog.}, 2016, pp.
  770--778.

\bibitem{liu2020learning}
N.~Liu, N.~Zhang, and J.~Han, ``Learning selective self-mutual attention for
  {RGB-D} saliency detection,'' in \emph{IEEE Conf. Comput. Vis. Pattern
  Recog.}, 2020, pp. 13\,756--13\,765.

\bibitem{hou2019deeply}
Q.~Hou, M.-M. Cheng, X.~Hu, A.~Borji, Z.~Tu, and P.~H. Torr, ``Deeply
  supervised salient object detection with short connections.'' \emph{IEEE
  Trans. Pattern Anal. Mach. Intell.}, vol.~41, no.~4, p. 815, 2019.

\bibitem{qin2019basnet}
X.~Qin, Z.~Zhang, C.~Huang, C.~Gao, M.~Dehghan, and M.~Jagersand, ``{BASNet}:
  Boundary-aware salient object detection,'' in \emph{IEEE Conf. Comput. Vis.
  Pattern Recog.}, 2019, pp. 7479--7489.

\bibitem{liu2019pool}
J.-J. Liu, Q.~Hou, M.-M. Cheng, J.~Feng, and J.~Jiang, ``A simple pooling-based
  design for real-time salient object detection,'' in \emph{IEEE Conf. Comput.
  Vis. Pattern Recog.}, 2019, pp. 3917--3926.

\bibitem{zhang2019salient}
P.~Zhang, W.~Liu, H.~Lu, and C.~Shen, ``Salient object detection with lossless
  feature reflection and weighted structural loss,'' \emph{IEEE Trans. Image
  Process.}, vol.~28, no.~6, pp. 3048--3060, 2019.

\bibitem{wu2021regularized}
Y.-H. Wu, Y.~Liu, L.~Zhang, W.~Gao, and M.-M. Cheng, ``Regularized
  densely-connected pyramid network for salient instance segmentation,''
  \emph{IEEE Trans. Image Process.}, vol.~30, pp. 3897--3907, 2021.

\bibitem{wu2021edn}
Y.-H. Wu, Y.~Liu, L.~Zhang, M.-M. Cheng, and B.~Ren, ``{EDN}: Salient object
  detection via extremely-downsampled network,'' \emph{arXiv preprint
  arXiv:2012.13093}, 2021.

\bibitem{yang2013saliency}
C.~Yang, L.~Zhang, H.~Lu, X.~Ruan, and M.-H. Yang, ``Saliency detection via
  graph-based manifold ranking,'' in \emph{IEEE Conf. Comput. Vis. Pattern
  Recog.}, 2013, pp. 3166--3173.

\bibitem{cheng2013efficient}
M.-M. Cheng, J.~Warrell, W.-Y. Lin, S.~Zheng, V.~Vineet, and N.~Crook,
  ``Efficient salient region detection with soft image abstraction,'' in
  \emph{Int. Conf. Comput. Vis.}, 2013, pp. 1529--1536.

\bibitem{cheng2014global}
M.-M. Cheng, N.~J. Mitra, X.~Huang, P.~H. Torr, and S.-M. Hu, ``Global contrast
  based salient region detection,'' \emph{IEEE Trans. Pattern Anal. Mach.
  Intell.}, vol.~37, no.~3, pp. 569--582, 2014.

\bibitem{wang2017salient}
J.~Wang, H.~Jiang, Z.~Yuan, M.-M. Cheng, X.~Hu, and N.~Zheng, ``Salient object
  detection: A discriminative regional feature integration approach,''
  \emph{Int. J. Comput. Vis.}, vol. 123, no.~2, pp. 251--268, 2017.

\bibitem{li2013saliency}
X.~Li, H.~Lu, L.~Zhang, X.~Ruan, and M.-H. Yang, ``Saliency detection via dense
  and sparse reconstruction,'' in \emph{Int. Conf. Comput. Vis.}, 2013, pp.
  2976--2983.

\bibitem{zhang2017saliency}
L.~Zhang, J.~Ai, B.~Jiang, H.~Lu, and X.~Li, ``Saliency detection via absorbing
  markov chain with learnt transition probability,'' \emph{IEEE Trans. Image
  Process.}, vol.~27, no.~2, pp. 987--998, 2017.

\bibitem{wu2019mutual}
R.~Wu, M.~Feng, W.~Guan, D.~Wang, H.~Lu, and E.~Ding, ``A mutual learning
  method for salient object detection with intertwined multi-supervision,'' in
  \emph{IEEE Conf. Comput. Vis. Pattern Recog.}, 2019, pp. 8150--8159.

\bibitem{liu2021part}
Y.~Liu, D.~Zhang, Q.~Zhang, and J.~Han, ``Part-object relational visual
  saliency,'' \emph{IEEE Trans. Pattern Anal. Mach. Intell.}, 2021.

\bibitem{han2018advanced}
J.~Han, D.~Zhang, G.~Cheng, N.~Liu, and D.~Xu, ``Advanced deep-learning
  techniques for salient and category-specific object detection: a survey,''
  \emph{IEEE Signal Processing Magazine}, vol.~35, no.~1, pp. 84--100, 2018.

\bibitem{borji2019salient}
A.~Borji, M.-M. Cheng, Q.~Hou, H.~Jiang, and J.~Li, ``Salient object detection:
  A survey,'' \emph{Computational Visual Media}, vol.~5, no.~2, pp. 117--150,
  2019.

\bibitem{wang2021salient}
W.~Wang, Q.~Lai, H.~Fu, J.~Shen, H.~Ling, and R.~Yang, ``Salient object
  detection in the deep learning era: An in-depth survey,'' \emph{IEEE Trans.
  Pattern Anal. Mach. Intell.}, 2021.

\bibitem{lang2012depth}
C.~Lang, T.~V. Nguyen, H.~Katti, K.~Yadati, M.~Kankanhalli, and S.~Yan, ``Depth
  matters: Influence of depth cues on visual saliency,'' in \emph{Eur. Conf.
  Comput. Vis.}, 2012, pp. 101--115.

\bibitem{ciptadi2013depth}
J.~R. Arridhana~Ciptadi, Tucker~Hermans, ``An in depth view of saliency,'' in
  \emph{Proceedings of the British Machine Vision Conference}.\hskip 1em plus
  0.5em minus 0.4em\relax BMVA Press, 2013.

\bibitem{desingh2013depth}
K.~Desingh, K.~M. Krishna, D.~Rajan, and C.~Jawahar, ``Depth really matters:
  Improving visual salient region detection with depth,'' in \emph{Proceedings
  of the British Machine Vision Conference}.\hskip 1em plus 0.5em minus
  0.4em\relax BMVA Press, 2013.

\bibitem{fan2014salient}
X.~Fan, Z.~Liu, and G.~Sun, ``Salient region detection for stereoscopic
  images,'' in \emph{International Conference on Digital Signal
  Processing}.\hskip 1em plus 0.5em minus 0.4em\relax IEEE, 2014, pp. 454--458.

\bibitem{peng2014rgbd}
H.~Peng, B.~Li, W.~Xiong, W.~Hu, and R.~Ji, ``{RGBD} salient object detection:
  a benchmark and algorithms,'' in \emph{Eur. Conf. Comput. Vis.}, 2014, pp.
  92--109.

\bibitem{cheng2014depth}
Y.~Cheng, H.~Fu, X.~Wei, J.~Xiao, and X.~Cao, ``Depth enhanced saliency
  detection method,'' in \emph{International Conference on Internet Multimedia
  Computing and Service}, 2014, pp. 23--27.

\bibitem{pang2020hierarchical}
Y.~Pang, L.~Zhang, X.~Zhao, and H.~Lu, ``Hierarchical dynamic filtering network
  for {RGB-D} salient object detection,'' in \emph{Eur. Conf. Comput. Vis.},
  2020, pp. 235--252.

\bibitem{zhao2021self}
X.~Zhao, Y.~Pang, L.~Zhang, H.~Lu, and X.~Ruan, ``Self-supervised
  representation learning for {RGB-D} salient object detection,'' \emph{arXiv
  preprint arXiv:2101.12482}, 2021.

\bibitem{chen2020progressively}
S.~Chen and Y.~Fu, ``Progressively guided alternate refinement network for
  {RGB-D} salient object detection,'' in \emph{Eur. Conf. Comput. Vis.}, 2020,
  pp. 520--538.

\bibitem{huang2020joint}
N.~Huang, Y.~Liu, Q.~Zhang, and J.~Han, ``Joint cross-modal and unimodal
  features for {RGB-D} salient object detection,'' \emph{IEEE Trans.
  Multimedia}, vol.~23, pp. 2428--2441, 2021.

\bibitem{qu2017rgbd}
L.~Qu, S.~He, J.~Zhang, J.~Tian, Y.~Tang, and Q.~Yang, ``{RGBD} salient object
  detection via deep fusion,'' \emph{IEEE Trans. Image Process.}, vol.~26,
  no.~5, pp. 2274--2285, 2017.

\bibitem{chen2021rgb}
Q.~Chen, Z.~Liu, Y.~Zhang, K.~Fu, Q.~Zhao, and H.~Du, ``{RGB-D} salient object
  detection via 3d convolutional neural networks,'' in \emph{AAAI Conf. Artif.
  Intell.}, 2021.

\bibitem{ji2021calibrated}
W.~Ji, J.~Li, S.~Yu, M.~Zhang, Y.~Piao, S.~Yao, Q.~Bi, K.~Ma, Y.~Zheng, H.~Lu
  \emph{et~al.}, ``Calibrated rgb-d salient object detection,'' in \emph{IEEE
  Conf. Comput. Vis. Pattern Recog.}, 2021, pp. 9471--9481.

\bibitem{zhou2021rgb}
T.~Zhou, D.-P. Fan, M.-M. Cheng, J.~Shen, and L.~Shao, ``{RGB-D} salient object
  detection: A survey,'' \emph{Computational Visual Media}, pp. 1--33, 2021.

\bibitem{han2017cnns}
J.~Han, H.~Chen, N.~Liu, C.~Yan, and X.~Li, ``{CNN}s-based {RGB-D} saliency
  detection via cross-view transfer and multiview fusion,'' \emph{IEEE Trans.
  Cybernetics}, vol.~48, no.~11, pp. 3171--3183, 2017.

\bibitem{wang2019adaptive}
N.~Wang and X.~Gong, ``Adaptive fusion for {RGB-D} salient object detection,''
  \emph{IEEE Access}, vol.~7, pp. 55\,277--55\,284, 2019.

\bibitem{huang2021employing}
N.~Huang, Y.~Yang, D.~Zhang, Q.~Zhang, and J.~Han, ``Employing bilinear fusion
  and saliency prior information for {RGB-D} salient object detection,''
  \emph{IEEE Trans. Multimedia}, 2021.

\bibitem{tan2019mnasnet}
M.~Tan, B.~Chen, R.~Pang, V.~Vasudevan, M.~Sandler, A.~Howard, and Q.~V. Le,
  ``Mnas{N}et: Platform-aware neural architecture search for mobile,'' in
  \emph{IEEE Conf. Comput. Vis. Pattern Recog.}, 2019, pp. 2820--2828.

\bibitem{mehta2018espnetv2}
S.~Mehta, M.~Rastegari, L.~Shapiro, and H.~Hajishirzi, ``{ESPNetv2}: A
  light-weight, power efficient, and general purpose convolutional neural
  network,'' in \emph{IEEE Conf. Comput. Vis. Pattern Recog.}, 2019, pp.
  9190--9200.

\bibitem{tan2020efficientdet}
M.~Tan, R.~Pang, and Q.~V. Le, ``Efficient{D}et: Scalable and efficient object
  detection,'' in \emph{IEEE Conf. Comput. Vis. Pattern Recog.}, 2020, pp.
  10\,781--10\,790.

\bibitem{liu2020lightweight}
Y.~Liu, Y.-C. Gu, X.-Y. Zhang, W.~Wang, and M.-M. Cheng, ``Lightweight salient
  object detection via hierarchical visual perception learning,'' \emph{IEEE
  Trans. Cybernetics}, vol.~51, no.~9, pp. 4439--4449, 2021.

\bibitem{liu2021samnet}
Y.~Liu, X.-Y. Zhang, J.-W. Bian, L.~Zhang, and M.-M. Cheng, ``{SAMN}et:
  Stereoscopically attentive multi-scale network for lightweight salient object
  detection,'' \emph{IEEE Trans. Image Process.}, vol.~30, pp. 3804--3814,
  2021.

\bibitem{ioffe2015batch}
S.~Ioffe and C.~Szegedy, ``Batch normalization: Accelerating deep network
  training by reducing internal covariate shift,'' in \emph{Int. Conf. Mach.
  Learn.}, 2015, pp. 448--456.

\bibitem{nair2010rectified}
V.~Nair and G.~E. Hinton, ``Rectified linear units improve restricted boltzmann
  machines,'' in \emph{Int. Conf. Mach. Learn.}\hskip 1em plus 0.5em minus
  0.4em\relax Madison, WI, USA: Omnipress, 2010, pp. 807--814.

\bibitem{wu2021jcs}
Y.-H. Wu, S.-H. Gao, J.~Mei, J.~Xu, D.-P. Fan, R.-G. Zhang, and M.-M. Cheng,
  ``{JCS}: An explainable covid-19 diagnosis system by joint classification and
  segmentation,'' \emph{IEEE Trans. Image Process.}, vol.~30, pp. 3113--3126,
  2021.

\bibitem{wang2004image}
Z.~Wang, A.~C. Bovik, H.~R. Sheikh, and E.~P. Simoncelli, ``Image quality
  assessment: from error visibility to structural similarity,'' \emph{IEEE
  Trans. Image Process.}, vol.~13, no.~4, pp. 600--612, 2004.

\bibitem{milletari2016v}
F.~Milletari, N.~Navab, and S.-A. Ahmadi, ``V-{N}et: Fully convolutional neural
  networks for volumetric medical image segmentation,'' in \emph{2016 fourth
  international conference on 3D vision (3DV)}.\hskip 1em plus 0.5em minus
  0.4em\relax IEEE, 2016, pp. 565--571.

\bibitem{cong2016saliency}
R.~Cong, J.~Lei, C.~Zhang, Q.~Huang, X.~Cao, and C.~Hou, ``Saliency detection
  for stereoscopic images based on depth confidence analysis and multiple cues
  fusion,'' \emph{IEEE Signal Processing Letters}, vol.~23, no.~6, pp.
  819--823, 2016.

\bibitem{chen2018progressively}
H.~Chen and Y.~Li, ``Progressively complementarity-aware fusion network for
  {RGB-D} salient object detection,'' in \emph{IEEE Conf. Comput. Vis. Pattern
  Recog.}, 2018, pp. 3051--3060.

\bibitem{zhang2021bilateral}
Z.~Zhang, Z.~Lin, J.~Xu, W.-D. Jin, S.-P. Lu, and D.-P. Fan, ``Bilateral
  attention network for rgb-d salient object detection,'' \emph{IEEE Trans.
  Image Process.}, vol.~30, pp. 1949--1961, 2021.

\bibitem{paszke2019pytorch}
A.~Paszke, S.~Gross, F.~Massa, A.~Lerer, J.~Bradbury, G.~Chanan, T.~Killeen,
  Z.~Lin, N.~Gimelshein, L.~Antiga \emph{et~al.}, ``Py{T}orch: An imperative
  style, high-performance deep learning library,'' in \emph{Annu. Conf. Neur.
  Inform. Process. Syst.}\hskip 1em plus 0.5em minus 0.4em\relax Curran
  Associates, Inc., 2019, pp. 8026--8037.

\bibitem{hu2020jittor}
S.-M. Hu, D.~Liang, G.-Y. Yang, G.-W. Yang, and W.-Y. Zhou, ``Jittor: a novel
  deep learning framework with meta-operators and unified graph execution,''
  \emph{Science China Information Sciences}, vol.~63, no.~12, pp. 1--21, 2020.

\bibitem{kingma2014adam}
D.~P. Kingma and J.~Ba, ``Adam: A method for stochastic optimization,'' in
  \emph{Int. Conf. Learn. Represent.}, 2015.

\bibitem{niu2012leveraging}
Y.~Niu, Y.~Geng, X.~Li, and F.~Liu, ``Leveraging stereopsis for saliency
  analysis,'' in \emph{IEEE Conf. Comput. Vis. Pattern Recog.}\hskip 1em plus
  0.5em minus 0.4em\relax IEEE, 2012, pp. 454--461.

\bibitem{zhu2017three}
C.~Zhu and G.~Li, ``A three-pathway psychobiological framework of salient
  object detection using stereoscopic technology,'' in \emph{Int. Conf. Comput.
  Vis. Worksh.}, 2017, pp. 3008--3014.

\bibitem{achanta2009frequency}
R.~Achanta, S.~Hemami, F.~Estrada, and S.~Susstrunk, ``Frequency-tuned salient
  region detection,'' in \emph{IEEE Conf. Comput. Vis. Pattern Recog.}, 2009,
  pp. 1597--1604.

\bibitem{fan2017structure}
D.-P. Fan, M.-M. Cheng, Y.~Liu, T.~Li, and A.~Borji, ``Structure-measure: A new
  way to evaluate foreground maps,'' in \emph{Int. Conf. Comput. Vis.}, 2017,
  pp. 4548--4557.

\bibitem{fan2018enhanced}
D.-P. Fan, G.-P. Ji, X.~Qin, and M.-M. Cheng, ``Cognitive vision inspired
  object segmentation metric and loss function,'' \emph{SCIENTIA SINICA
  Informationis}, vol.~6, 2021.

\bibitem{chen2021spatial}
L.-Z. Chen, Z.~Lin, Z.~Wang, Y.-L. Yang, and M.-M. Cheng, ``Spatial information
  guided convolution for real-time {RGBD} semantic segmentation,'' \emph{IEEE
  Trans. Image Process.}, vol.~30, pp. 2313--2324, 2021.

\bibitem{chen2020bi}
X.~Chen, K.-Y. Lin, J.~Wang, W.~Wu, C.~Qian, H.~Li, and G.~Zeng,
  ``Bi-directional cross-modality feature propagation with
  separation-and-aggregation gate for {RGB-D} semantic segmentation,'' in
  \emph{Eur. Conf. Comput. Vis.}, 2020, pp. 561--577.

\bibitem{nyudv2}
N.~Silberman, D.~Hoiem, P.~Kohli, and R.~Fergus, ``Indoor segmentation and
  support inference from {RGBD} images,'' in \emph{Eur. Conf. Comput. Vis.},
  2012, pp. 746--760.

\bibitem{wu2021p2t}
Y.-H. Wu, Y.~Liu, X.~Zhan, and M.-M. Cheng, ``{P2T}: Pyramid pooling
  transformer for scene understanding,'' \emph{arXiv preprint
  arXiv:2106.12011}, 2021.

\end{thebibliography}
}

\newcommand{\AddPhoto}[1]{\includegraphics[width=1in,keepaspectratio]{#1}}

\begin{IEEEbiography}[\AddPhoto{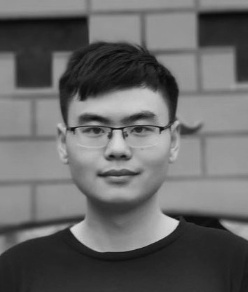}]{Yu-Huan Wu}
 is currently a Ph.D. candidate with College of Computer Science 
 at Nankai University, supervised by Prof. Ming-Ming Cheng. 
 He received his bachelor's degree from Xidian University in 2018. 
 His research interests include computer vision
 and machine learning.
\end{IEEEbiography}

\begin{IEEEbiography}[\AddPhoto{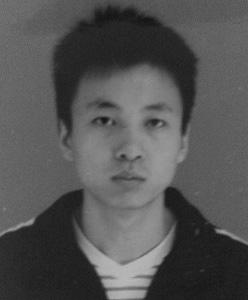}]{Yun Liu}
 received his bachelor's and doctoral degrees from Nankai University
 in 2016 and 2020, respectively.
 Currently, he works as a postdoctoral scholar with Prof. Luc Van Gool 
 at ETH Zurich.
 His research interests include computer vision and machine learning.
\end{IEEEbiography}

\begin{IEEEbiography}[\AddPhoto{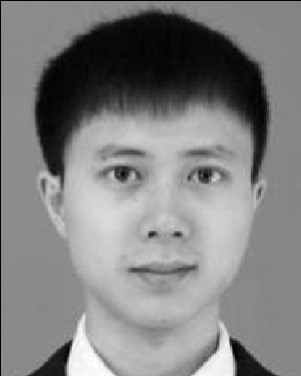}]{Jun Xu}
received his B.Sc. and M.Sc. degrees from School of Mathematics Science, Nankai University, Tianjin, China, in 2011 and 2014, respectively, and the Ph.D. degree from the Department of Computing, Hong Kong Polytechnic University, in 2018. He worked as a Research Scientist at IIAI, Abu Dhabi, UAE. He is currently a Lecturer with School of Statistics and Data Science, Nankai University. More information can be found at \url{https://csjunxu.github.io/}.
\end{IEEEbiography}

\begin{IEEEbiography}[\AddPhoto{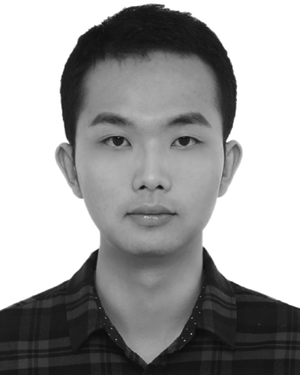}]{Jia-Wang Bian}
Jia-Wang Bian received the B.Eng. degree from Nankai University, where he was advised by Prof. M.-M. Cheng. He is currently pursuing the Ph.D. degree with The University of Adelaide. He was a Research Assistant with the Singapore University of Technology and Design (SUTD). He also did a Trainee Engineer Job with the Advanced Digital Sciences Center (ADSC), Huawei Technologies Co., Ltd., and Tusimple. He is also an Associated Ph.D. Researcher with the Australian Centre for Robotic Vision (ACRV). He is advised by Prof. I. Reid and Prof. C. Shen. His research interests include computer vision and robotics.
\end{IEEEbiography}

\begin{IEEEbiography}[\AddPhoto{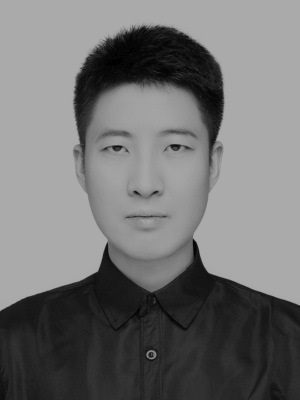}]{Yu-Chao Gu}
 received his bachelor’s degree from the Beijing University of Chemical Technology in 2019. 
He is currently pursuing the master's degree with the College of Computer Science, Nankai University.
His research interests include efficient deep-learning and computer vision.
\end{IEEEbiography}

\begin{IEEEbiography}[\AddPhoto{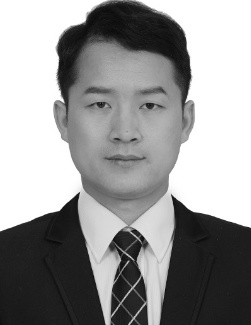}]{Ming-Ming Cheng}
 received his PhD degree from Tsinghua University in 2012.
 Then he did two years research fellow with Prof. Philip Torr
 in Oxford.
 He is now a professor at Nankai University, leading the
 Media Computing Lab.
 His research interests include computer graphics, computer
 vision, and image processing.
 He received research awards, including ACM China Rising Star Award,
 IBM Global SUR Award, and CCF-Intel Young Faculty Researcher Program.
 He is on the editorial boards of IEEE TPAMI and IEEE TIP.
\end{IEEEbiography}

\vfill

\end{document}